\documentclass{article} 

\usepackage{preprint,times}
\iclrpreprintcopy

\usepackage{amsmath,amsfonts,bm}

\def\eqref#1{equation~\ref{#1}}

\def\1{\bm{1}}

\DeclareMathAlphabet{\mathsfit}{\encodingdefault}{\sfdefault}{m}{sl}
\SetMathAlphabet{\mathsfit}{bold}{\encodingdefault}{\sfdefault}{bx}{n}

\usepackage[utf8]{inputenc} 
\usepackage[T1]{fontenc}    
\usepackage{hyperref}       
\usepackage{url}            
\usepackage{booktabs}       
\usepackage{amsfonts}       
\usepackage{nicefrac}       
\usepackage{microtype}      
\usepackage{hyperref} 

\usepackage{booktabs}
\usepackage{adjustbox}
\usepackage{pdflscape}
\usepackage{rotating}

\usepackage{amsmath}
\usepackage{amssymb}
\usepackage{graphicx}
\usepackage{wrapfig}
\usepackage{sidecap}
\usepackage{algorithm2e}
\usepackage{multirow}
\RestyleAlgo{ruled}
\LinesNumbered
\DontPrintSemicolon
\usepackage[english]{babel}

\usepackage{float}
\usepackage{enumitem}
\usepackage{titlesec}
\usepackage{setspace}
\usepackage{enumitem}
\usepackage[table]{xcolor}
\usepackage{booktabs}
\setlist[itemize]{leftmargin=*}

\usepackage{amsthm}

\definecolor{first}{HTML}{D9EAD3}   
\definecolor{second}{HTML}{D9EAF7}  
\definecolor{third}{HTML}{FFF2CC}   

\hypersetup{hidelinks}

\author{
\textbf{Quanling Zhao}$^1$ \quad
\textbf{Jiaying Yang}$^2$ \quad
\textbf{Tianqi Zhang}$^1$ \quad
\textbf{Ziyang Hao}$^1$ \\
\textbf{Fatemeh Asgarinejad}$^3$ \quad
\textbf{Flavio Ponzina}$^4$ \quad
\textbf{Tajana Rosing}$^1$ \\
$^1$University of California San Diego \quad $^2$ Georgia Institute of Technology\\
$^3$ University of California Riverside \quad $^4$ San Diego State University\\
\texttt{\{quzhao,tiz014,z2hao,tajana\}@ucsd.edu} \quad \texttt{jesyjy7@gatech.edu} \\
\texttt{fatemeh.asgarinejad1@ucr.edu} \quad \texttt{fponzina@sdsu.edu}
}

\begin{document}

\title{Superposed Latent Autoencoder}

\maketitle

\begin{abstract}

Autoencoders typically meet tight latent-memory budgets by making each latent representation smaller, sacrificing representational capacity. We ask a different question: can multiple wider latents be stored together instead? We introduce the \textbf{Superposed Latent Autoencoder (SLAE)}, which preserves high-capacity latent representations while sharing storage through learned superposition. SLAE transforms latents into storage-friendly codes, binds them with randomized keys, superposes multiple codes into a single memory tensor, and learns to recover each latent before decoding. Under the same storage budget, SLAE replaces irreversible dimensional bottlenecks with structured interference that can be suppressed. Across CIFAR-10/100, SVHN, STL-10, Tiny ImageNet, and a wide range of memory budgets, SLAE substantially improves the reconstruction--memory tradeoff, reducing reconstruction error by up to 56\% over conventional autoencoders at matched storage. Further analysis shows that SLAE's advantage comes from making wider representations usable under the same storage budget. These gains also extend beyond reconstruction: the information preserved by SLAE improves downstream classification by up to 16.79 percentage points under the same memory budget. Our results suggest a new principle for representation compression: instead of making every latent smaller, keep representations wide and let them share memory.

\end{abstract}

\section{Introduction}

Autoencoders compress inputs into latent representations that retain the information needed for reconstruction and other downstream tasks~\citep{hinton2006reducing,bengio2013representation,rannen2017encoder}. Under a limited storage budget, the standard solution is to make these representations smaller by reducing their dimensionality or spatial resolution. This saves memory, but at the cost of representational capacity~\citep{wu2025h3ae,chen2024compressing,hinton2006reducing}.
\begin{wrapfigure}{r}{0.6\columnwidth}
    \centering
    \vspace{-4pt}
    \includegraphics[width=\linewidth]{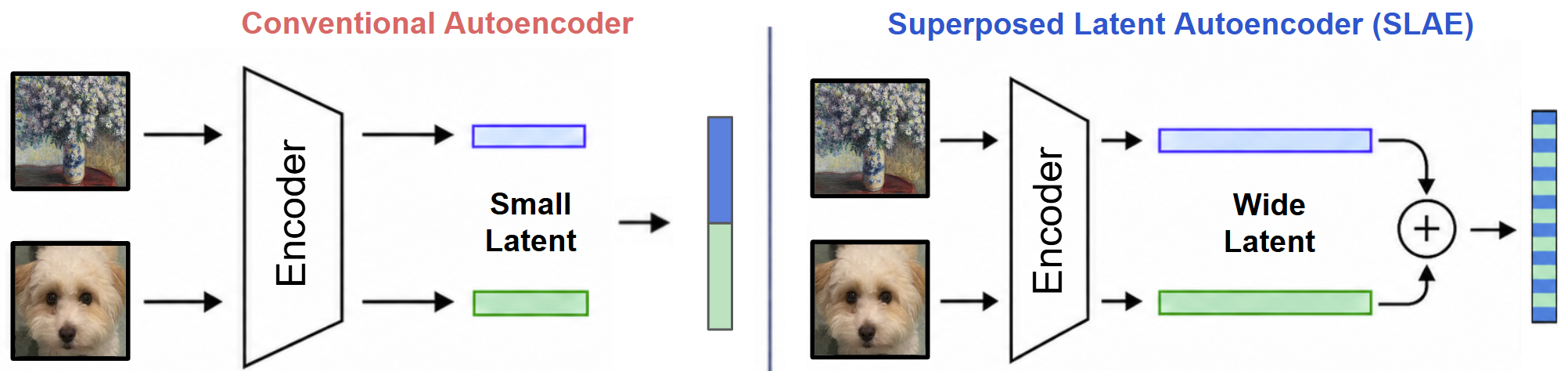}
    \vspace{-7mm}
    \caption{\small
    \textbf{AE vs.\ SLAE at matched storage.}
    SLAE shares wider latent representations through superposition instead
    of shrinking each latent independently.}
    \label{fig:concept}
    \vspace{-8pt}
\end{wrapfigure}
As the latent shrinks, increasingly more information must be discarded, turning memory compression into an increasingly severe information bottleneck. This raises a natural question: {does saving memory necessarily require making every representation smaller?} We explore a different strategy. Instead of storing each example independently in a small latent, we encode each example into a wider latent and allow several such latents to share the same memory through superposition. The shared memory is larger than a single conventional latent, but because it stores multiple examples at once, the average storage per example remains unchanged. This gives each example access to a richer representation without increasing its storage cost. The tradeoff is that the superposed latents interfere with one another, so the individual representations must be recovered from their shared memory~\citep{plate1995holographic,cheung2019superposition,menet2023mimonets}.

We introduce the \textbf{Superposed Latent Autoencoder (SLAE)}, which learns to make this shared-storage strategy work. Instead of storing each latent independently, SLAE first encodes each input into a wider latent representation. It then transforms the latent into a storage-friendly code, binds each code with a key, and superposes multiple codes into a single shared memory tensor~\citep{kanerva2009hyperdimensional,plate1995holographic}. To retrieve an individual representation, SLAE demixes the shared memory using the corresponding key and applies a learned recovery network before decoding. In this way, multiple high-capacity latent representations occupy the same physical storage.

The key idea behind SLAE is that a memory constraint creates a choice between two different kinds of information loss. A conventional autoencoder reduces latent dimensionality and directly sacrifices representational capacity~\citep{theis2017lossy,chen2025deep}. Instead, SLAE keeps a wider representation and pays for it through interference introduced by superposition. Randomized binding makes this interference structured~\citep{kleyko2022vector}, while the storage and recovery networks are trained jointly with the autoencoder to suppress its effect. SLAE therefore replaces an irreversible capacity bottleneck with a capacity--interference tradeoff. When the information gained from a wider latent exceeds the residual interference left after recovery, sharing storage can be preferable to shrinking each representation independently. This view also gives an intuitive prediction about when superposition should be useful. Under a very tight storage budget, a conventional latent may be severely capacity limited, so increasing its width can recover substantial information. In this regime, SLAE can tolerate a meaningful amount of interference and still provide a net benefit. As the storage budget grows and the conventional latent becomes more expressive, the benefit of additional width naturally becomes smaller. We therefore expect the strongest gains from SLAE in regimes where latent capacity is the primary bottleneck.

Our experiments support this picture. We evaluate SLAE on image datasets across multiple latent geometries, storage budgets, and superposition factors. At matched average storage, SLAE substantially improves the reconstruction--memory tradeoff in constrained regimes, reducing reconstruction error by up to \textbf{56\%} compared with conventional autoencoders. Further analysis shows that SLAE's advantage comes from making wider representations usable under the same storage budget. The benefit also extends beyond reconstruction: at the same memory budget, representations preserved by SLAE improve downstream classification accuracy by up to \textbf{16.79} percentage points. Together, these results show that superposition provides a practical alternative to reducing latent dimensionality when representation storage is limited. Overall, our contributions are threefold:
\begin{itemize}[nosep]
    \item We introduce \textbf{Superposed Latent Autoencoder (SLAE)}, a new architecture to compress latent representations by storing multiple wide latents together rather than shrinking each one independently.
    
    \item We formulate superposed latent storage as a capacity--interference tradeoff: rather than irreversibly discarding information through smaller latents, SLAE preserves latent capacity while learning to suppress the interference introduced by shared storage.
    
    \item We demonstrate the effectiveness of SLAE across five image datasets spanning different sizes and resolutions. At matched storage, SLAE reduces reconstruction error by up to \textbf{56\%} and improves downstream classification by up to \textbf{16.79} percentage points.
\end{itemize}

\section{Related Work}

\textbf{Autoencoders and latent compression: } Autoencoders compress data by mapping inputs through a restricted latent representation, making latent capacity central to reconstruction quality. Recent work has studied this tradeoff from several directions. Least-Volume autoencoders explicitly reduce the effective dimensionality of the latent space~\citep{chen2024compressing}, while recent theory characterizes how autoencoder architecture and data structure affect compression quality~\citep{cosentino2022deep,psenka2024representation,chakraborty2024statistical}. Learned image compression further improves rate--distortion performance through learned transforms, quantization, entropy models, and more expressive latent representations~\citep{balle2018variational,he2022elic,balle2016end}. Despite these advances, the storage budget is primarily addressed by making each example's representation individually more compact. More broadly, recent work has also compressed stored learned representations, including neural features and language-model KV states~\citep{li2024nerfcodec,dong2024get}. These methods further motivate representation memory as an important resource, but still compress individual representations rather than sharing storage across multiple wider latents. SLAE considers a different axis: rather than further shrinking each latent, it preserves wider representations and amortizes their storage across multiple examples.

\textbf{Information superposition: } Superposition originates from distributed representation frameworks such as holographic reduced representations and vector-symbolic architectures~\citep{plate1995holographic,kanerva2009hyperdimensional}. Superposition has since been used to store multiple neural networks within a shared parameter tensor~\citep{cheung2019superposition}, and more recently studied as a form of lossy compression for overlapping neural features~\citep{bereska2025superposition}. Related neural multiplexing methods such as DataMUX~\citep{murahari2022datamux} combine multiple transformed inputs into a shared representation for joint neural processing, while MIMONets~\citep{menet2023mimonets} use binding and superposition to process multiple inputs through CNNs and Transformers before unbinding their outputs. These approaches primarily use superposition to amortize computation and improve inference throughput. SLAE instead uses superposition to amortize persistent latent storage: each example retains a wider autoencoder representation, while multiple such representations share the same physical memory under a fixed average storage budget. SLAE therefore changes the unit of latent compression from an individually compressed representation to a group of jointly stored representations.

\section{Motivation and Problem Formulation}
\label{sec:motivation}

\begin{wrapfigure}{r}{0.4\columnwidth}
    \centering
    \vspace{-12pt}
    \includegraphics[width=\linewidth]{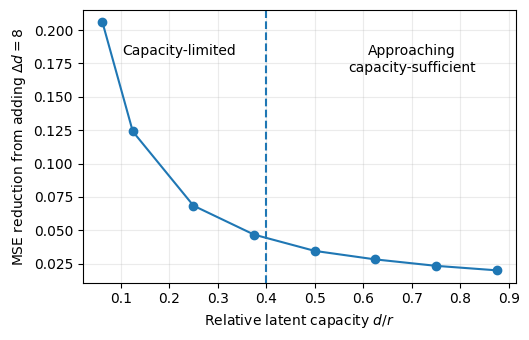}
    \vspace{-8mm}
    \caption{\small Adding latent width ($\Delta d=8$) yields the largest reconstruction gain when the autoencoder is capacity limited, with diminishing benefit as $d/r$ grows. The dashed line qualitatively separates the two regimes.}
    \label{fig:capacity_motivation}
    \vspace{-10pt}
\end{wrapfigure}
\textbf{A motivating capacity effect: } The value of additional autoencoder latent capacity depends strongly on the bottleneck regime. To isolate this effect, we consider a linear autoencoder on synthetic vector data with $r$ signal-bearing directions whose importance decays smoothly, and vary its latent dimension $d$~\citep{baldi1989neural}. Figure~\ref{fig:capacity_motivation} shows the reduction in reconstruction error obtained by adding latent dimensions. When the autoencoder latent is strongly capacity limited, additional dimensions provide large reconstruction gains; as the latent becomes more expressive, the same increase in width becomes progressively less valuable. This motivates the regime targeted by SLAE: under tight storage budgets, where a conventional autoencoder is forced to use a small latent, preserving additional latent capacity can be particularly valuable. More broadly, this capacity limitation is fundamental to bottlenecked autoencoders rather than specific to a particular architecture~\citep{bourlard1988auto,baldi1989neural,cosentino2022deep}, making effective use of limited latent capacity an important problem in representation compression.

\textbf{Matched-storage formulation: } Given a fixed per-example storage budget $B$, our goal is to minimize reconstruction error while storing no more than $B$ scalars per example. A conventional autoencoder encodes an input $x$ into a spatial latent $z\in\mathbb{R}^{C\times H\times H}$~\citep{masci2011stacked,balle2016end}, where $C$ denotes the latent channel dimension and $H$ the spatial size of the square latent feature map, with $B=CH^2$ stored scalars per example. For $K$ examples, this requires $KB$ stored scalars in total. We ask whether the same storage constraint can support higher-capacity representations by allowing examples to share memory. Rather than restricting each example to a $B$-scalar latent, SLAE allows each example to use a wider representation $z_k\in\mathbb{R}^{KC\times H\times H}$ containing $KB$ scalars. These wider latents are not stored independently. Instead, $K$ of them are transformed and superposed into a single shared memory $m\in\mathbb{R}^{KC\times H\times H}$. The average physical storage per example is therefore $|m|/K=(KCH^2)/K=CH^2=B$, exactly matching the conventional autoencoder.

The two strategies therefore make different compromises under the same memory budget. A conventional autoencoder stores each example independently but may force its latent into a capacity-limited regime. SLAE instead gives each example access to a wider representation while introducing interference through shared storage. The central problem is whether the information preserved by the wider latent can outweigh the residual interference after recovery, yielding lower reconstruction error under the same storage constraint. When latent capacity is the dominant bottleneck, even imperfect recovery of a wider representation can be preferable to irreversibly shrinking each latent independently. We focus on convolutional image autoencoders for the remainder of the paper as a clear and controlled setting for studying the reconstruction--memory tradeoff. The underlying principle is more general: SLAE is not inherently image-specific and can in principle be applied to other latent representations.

\section{Superposed Latent Autoencoder}
\label{sec:method}

\begin{figure}[t]
    \centering
    \includegraphics[width=0.8\textwidth]{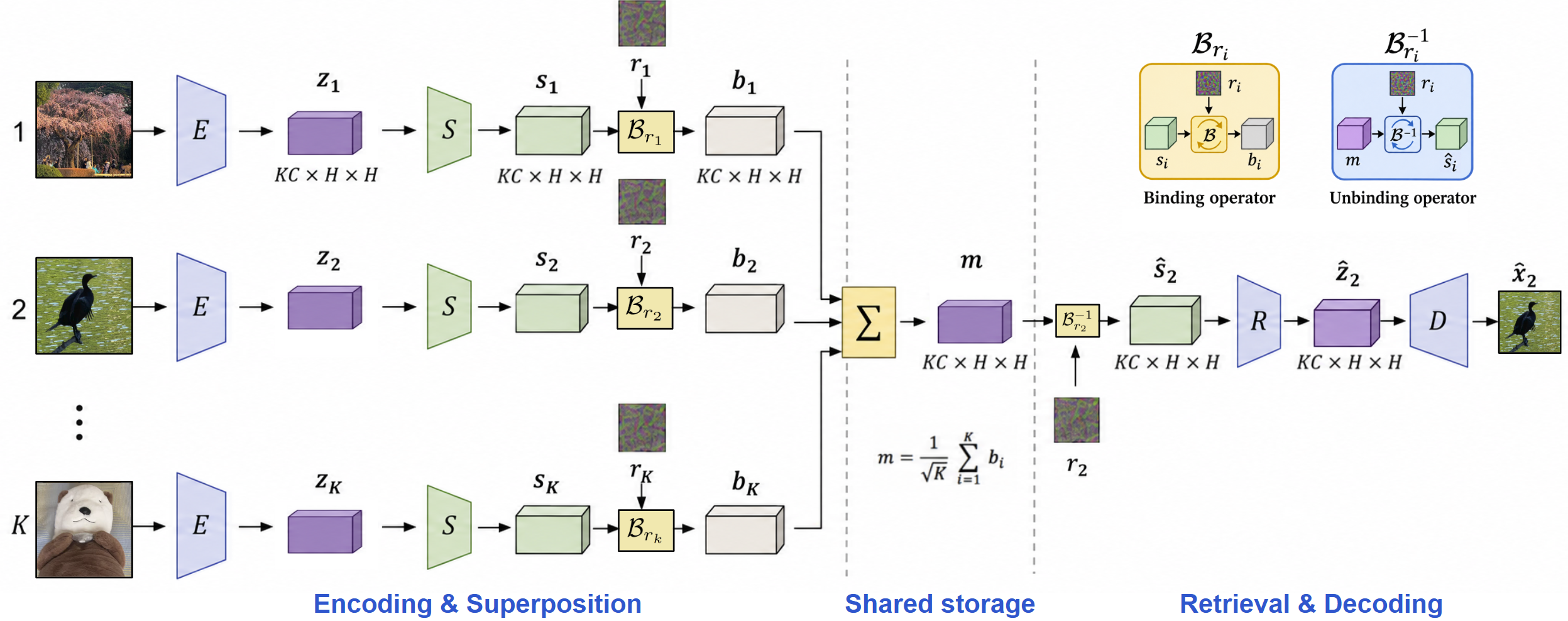}
    \vspace{-4mm}
\caption{\textbf{SLAE Architecture.} Each input $x_i$ is encoded by $E$ into a wide latent $z_i$, transformed by the storage adapter $S$ into $s_i$, bound with a slot-specific key $r_i$, and superposed with the other codes into shared memory $m$. Retrieval applies the corresponding inverse binding to obtain $\hat{s}_i$, which the recovery network $R$ maps to a decoder-friendly latent $\hat{z}_i$ for reconstruction by $D$. SLAE thus enables multiple wide latents to share the same memory through learned storage and recovery.}
\label{fig:slae_method}
\end{figure}

The central design of SLAE is to decouple the representation used for reconstruction from the representation used for storage. The encoder is free to produce a high-capacity latent suited to the decoder, while a separate learned pathway adapts that latent for superposition and later recovers it for decoding. This separation allows SLAE to optimize for shared storage without forcing the autoencoder latent itself to be directly superposition-friendly.

\subsection{Method Overview}

Consider a group of $K$ inputs $\{x_i\}_{i=1}^{K}$. As illustrated in Figure~\ref{fig:slae_method}, each input is first encoded by the encoder $E$ into a wide latent representation $z_i=E(x_i)$. A learned storage adapter $S$ then transforms $z_i$ into a storage-friendly code $s_i$. Each $s_i$ is bound with a slot-specific randomized key $r_i$ through an invertible binding operator $\mathcal{B}_{r_i}$, producing a bound representation $b_i$. The $K$ bound representations are then superposed into a single shared memory tensor $m$. To retrieve example $k$, SLAE applies the corresponding inverse binding operator $\mathcal{B}_{r_k}^{-1}$ to the shared memory, producing a demixed storage code $\hat{s}_k$. Because all $K$ representations occupy the same memory, $\hat{s}_k$ contains the desired code together with interference from the other stored examples. A learned recovery network $R$ maps this demixed representation back to a decoder-friendly latent $\hat{z}_k$, which is reconstructed by the decoder $D$ as $\hat{x}_k=D(\hat{z}_k)$. The resulting pipeline preserves the familiar encoder--decoder structure of an autoencoder while introducing learned storage and recovery specifically for shared latent storage.

\subsection{Shared Latent Storage}

\textbf{Storage-friendly latent codes: } SLAE operates on groups of $K$ examples that share one memory tensor. In all experiments, examples are randomly partitioned into groups of $K$ samples. Under the matched-storage setting of Section~\ref{sec:motivation}, SLAE uses a wider latent $z_i=E(x_i)\in\mathbb{R}^{KC\times H\times H}$, whereas a conventional autoencoder with the same average storage budget uses only $C\times H\times H$ values per example. Directly superposing these latents, however, is not ideal: the representation preferred by the decoder is not necessarily the representation best suited for shared storage. We therefore introduce a learned storage adapter $S$, which produces a storage-friendly code $s_i=S(z_i)\in\mathbb{R}^{KC\times H\times H}$ without changing its size. In our implementation, $S$ is a shallow residual convolutional adapter~\citep{he2016deep} that lightly transforms the encoder latent while preserving its spatial structure. It is identity-initialized, so it starts from the original autoencoder latent and learns only the adjustment needed for effective superposition and recovery.

\textbf{Randomized binding: } To distinguish the $K$ representations occupying the same memory, each slot $i$ is assigned a fixed randomized key $r_i$, represented by a random channel permutation $P_i$ and random sign pattern $D_i$. The key specifies an orthogonal binding operator $\mathcal{B}_{r_i}$, which transforms the storage code as $b_i=\mathcal{B}_{r_i}(s_i)=\mathcal{H}D_iP_i s_i$, where $\mathcal{H}$ is a normalized Walsh--Hadamard transform. Concretely, at each spatial location, the channel vector of $s_i$ is permuted, sign-flipped, and then mixed by the Hadamard transform. The same operation is applied independently across all $H\times H$ locations, preserving the spatial organization of the latent. Since $P_i$, $D_i$, and $\mathcal{H}$ are orthogonal, binding is exactly invertible, with $\mathcal{B}_{r_i}^{-1}=P_i^{-1}D_i\mathcal{H}$. For each slot, $P_i$ is sampled as a random permutation of the channel indices, while $D_i$ is a diagonal matrix with independent Rademacher entries in $\{-1,+1\}$. The normalized Walsh--Hadamard transform $\mathcal{H}$ is fixed and shared across slots.

Randomized orthogonal binding follows the intuition of parameter superposition~\citep{cheung2019superposition}. Before addition, each slot is placed in a different randomized orientation of the latent space. This preserves the information and norm of each code while making other slots appear as weakly aligned interference after unbinding. In our design, the random signed permutation $D_iP_i$ provides the slot-specific randomization, while the shared normalized Walsh--Hadamard transform $\mathcal{H}$ provides a fast, norm-preserving dense mixing of the memory representation. Because $\mathcal{H}$ is shared across slots, it cancels during unbinding and therefore does not itself contribute to cross-slot separation. Key overhead is small, each slot stores only a permutation and sign pattern over the $KC$ channels. These keys are fixed globally and reused across all superposition groups, so their cost does not scale with the number of stored examples. The shared Hadamard transform is fixed and applied implicitly using the Fast Walsh--Hadamard Transform rather than stored as a dense matrix~\citep{ailon2006approximate}.

\textbf{Superposition: } The $K$ bound representations are combined by element-wise addition into a single shared memory tensor:
\begin{equation}
\small
m=\frac{1}{\sqrt{K}}\sum_{i=1}^{K}b_i
=\frac{1}{\sqrt{K}}\sum_{i=1}^{K}\mathcal{B}_{r_i}(s_i),
\qquad
m\in\mathbb{R}^{KC\times H\times H}
\end{equation}

Since the binding transforms are orthogonal, $\|b_i\|_2=\|s_i\|_2$. For fixed storage codes and independently sampled keys, cross-slot inner products vanish in expectation, $\mathbb{E}_r\langle b_i,b_j\rangle=0$ for $i\neq j$. Therefore, $\mathbb{E}_r\|m\|_2^2=\frac{1}{K}\sum_{i=1}^{K}\|s_i\|_2^2$. Thus, the factor $1/\sqrt{K}$ keeps the energy of the shared memory at the average energy scale of its constituent codes, preventing a trivial $K$-fold increase due to summation and allowing the recovery network to focus on cross-slot interference. Importantly, only $m$ is stored, and $z_i$, $s_i$ are intermediate representations. Hence, the average storage per example is:
\begin{equation}
\small
\frac{|m|}{K}=\frac{KCH^2}{K}=CH^2=B
\end{equation}
exactly matching a conventional non-superposed autoencoder with a narrower latent. SLAE therefore provides each example with a wider latent representation at the same average storage cost, in exchange for interference introduced by superposition.

\subsection{Latent Recovery and Training}

\textbf{Demixing and recovery: } To retrieve example $k$, SLAE applies the inverse binding transform associated with its key to the shared memory. Accounting for the normalization used during superposition, the demixed storage code is
\begin{equation}
\small
\hat{s}_k
=
\sqrt{K}\,\mathcal{B}_{r_k}^{-1}(m)
=
s_k+\sum_{i\neq k}\mathcal{B}_{r_k}^{-1}\mathcal{B}_{r_i}(s_i)
\end{equation}
The desired storage code appears exactly as the first term, while the remaining terms represent interference from the other $K-1$ slots. Randomized binding makes these interfering representations weakly aligned with the target but does not remove them completely. SLAE therefore introduces a learned recovery network $R$ that maps the demixed storage code back to the decoder-friendly latent space. We write $\hat{z}_k=R(\hat{s}_k;k)$ and reconstruct the input as $\hat{x}_k=D(\hat{z}_k)$. The recovery network is an identity-initialized shallow residual network that combines local convolutional processing with lightweight mixing across spatial latent positions. It is conditioned on the retrieved slot $k$ using FiLM-style modulation~\citep{perez2018film} where a learned slot embedding produces channel-wise scale and shift parameters that modulate normalized features within the recovery blocks, allowing $R$ to adapt to the slot-specific interference pattern induced by the randomized binding keys.

\textbf{Training: }We initialize $E$ and $D$ from a conventional autoencoder trained using the same wide latent geometry as SLAE, while $S$ and $R$ are identity-initialized. This initialization provides SLAE with a high-capacity reconstruction latent before learning how to store and recover it through superposition. We first train the clean pathway $E\!\rightarrow\!S\!\rightarrow\!R\!\rightarrow\!D$ without superposition for a short warmup period, and then optimize the complete $K$-way superposed pathway end-to-end.

The training objective combines superposed reconstruction, latent recovery, a small clean-path reconstruction anchor, and a weak storage-code decorrelation penalty:
\begin{equation}
\small
\mathcal{L}
=
\lambda_x\mathcal{L}_{\mathrm{sup}}
+
\lambda_z\mathcal{L}_{\mathrm{latent}}
+
\lambda_{\mathrm{clean}}\mathcal{L}_{\mathrm{clean}}
+
\lambda_{\mathrm{decor}}\mathcal{L}_{\mathrm{decor}}
\end{equation}

Here, $\mathcal{L}_{\mathrm{sup}}
=
\frac{1}{K}\sum_{k=1}^{K}
\ell_{\mathrm{rec}}(\hat{x}_k,x_k)$
is the reconstruction loss for the superposed pathway, while
$\mathcal{L}_{\mathrm{latent}}
=
\frac{1}{K}\sum_{k=1}^{K}
\mathrm{MSE}(\hat{z}_k,z_k)$
directly encourages recovery of the original decoder-friendly latent. The reconstruction objective $\ell_{\mathrm{rec}}$ combines pixel-wise $\ell_1$ error, mean-squared error, and an SSIM term~\citep{wang2004image}. $\mathcal{L}_{\mathrm{clean}}$ applies the same reconstruction objective to the interference-free pathway (without superposition) $E\!\rightarrow\!S\!\rightarrow\!R\!\rightarrow\!D$. This term acts as a small anchor to the underlying autoencoder representation, discouraging the model from drifting toward a solution that is overly specialized to superposition interference at the expense of clean reconstruction. $\mathcal{L}_{\mathrm{decor}}$ regularizes the storage representation by penalizing correlations between different channels of $s_i$, following the
general idea of representation decorrelation~\citep{cogswell2015reducing,zbontar2021barlow}. Specifically, it penalizes the squared off-diagonal entries of the channel correlation matrix, encouraging the storage adapter to produce less redundant codes that are better suited for randomized superposition. After the warmup, $E$, $S$, $R$, and $D$ are optimized jointly under the full objective, allowing the encoder, storage code, recovery process, and decoder to co-adapt to the interference introduced by shared storage.

\section{Understanding Latent Superposition}
\label{sec:analysis}

\textbf{Effective interference under randomized binding: } We first examine what remains after unbinding a superposed representation. Retrieving slot $k$ applies the corresponding inverse binding transform and restores the original scale,
$\hat{s}_k=\sqrt{K}\,\mathcal{B}_{r_k}^{-1}(m)=s_k+\eta_k,\eta_k=\sum_{i\neq k}Q_{ki}s_i$, where $Q_{ki}=\mathcal{B}_{r_k}^{-1}\mathcal{B}_{r_i}$.
Thus, the desired storage code $s_k$ is recovered exactly before learned recovery, while the other $K-1$ codes remain as cross-slot interference.

Exact recovery cannot be expected for arbitrary storage codes. If each storage tensor were flattened into an unconstrained vector $s_i\in\mathbb{R}^{D}$, where $D=KCH^2$, then the $K$ codes would contain $KD$ degrees of freedom while the shared memory contains only $D$. The superposition map is therefore non-invertible in general. The relevant question is whether randomized binding structures the resulting interference so that useful information can nevertheless remain accessible after superposition. At first sight, the interference appears severe. Each relative binding transform $Q_{ki}$ is orthogonal and therefore preserves the energy of the interfering code: $\|Q_{ki}s_i\|_F=\|s_i\|_F$. For fixed storage codes and independently sampled slot keys, cross-slot terms vanish in expectation, giving $\mathbb{E}_{r}\|\eta_k\|_F^2=\sum_{i\neq k}\|s_i\|_F^2$.
Randomized binding therefore does not make the interference small in norm, the aggregate interference can have energy comparable to, or even larger than the desired code.

Following the intuition used in parameter superposition~\citep{cheung2019superposition}, the key observation is that preserving the total interference energy does not imply equally strong interference along any particular feature direction. Consider one spatial location $(h,w)$, where binding acts over the $d=KC$ channel dimensions. Let $v_i=s_i[:,h,w]\in\mathbb{R}^{d}$ denote the channel vector of storage code $s_i$ at that location. After unbinding, the corresponding channel vector is
$\hat{v}_k=v_k+\xi_k^{(h,w)}$, where
$\xi_k^{(h,w)}=\sum_{i\neq k}Q_{ki}v_i$ is the interference at location $(h,w)$. Let $a\in\mathbb{R}^{d}$ be any fixed linear probe independent of the sampled keys. Applying this probe gives
$a^\top\hat{v}_k=a^\top v_k+a^\top\xi_k^{(h,w)}$.
We define the scalar interference observed by this feature computation as:
\begin{equation}
\small
\epsilon_k^{(h,w)}
\triangleq
a^\top\xi_k^{(h,w)}
=
\sum_{i\neq k}a^\top Q_{ki}v_i
\end{equation}
Thus, $\xi_k^{(h,w)}$ denotes its channel vector at spatial location $(h,w)$, and $\epsilon_k^{(h,w)}$ denotes the scalar interference seen by a particular direction $a$. For brevity, we omit the spatial superscript $(h,w)$ in the following analysis. For the randomized binding used by SLAE, and for any fixed vectors $a,v\in\mathbb{R}^{d}$:
\begin{equation}
\label{eq:interference}
\small
\mathbb{E}_{r}[a^\top Qv]=0,
\qquad
\mathbb{E}_{r}\!\left[(a^\top Qv)^2\right]
=
\frac{\|a\|_2^2\|v\|_2^2}{d}
\end{equation}
Applying the first relation to all $K-1$ interfering slots gives
$\mathbb{E}_{r}[\epsilon_k]=0$.
Thus, randomized cross-slot interference has no systematic bias along any fixed feature direction. Its magnitude is characterized by its variance. Since $\epsilon_k$ is zero-mean,
$\operatorname{Var}_{r}(\epsilon_k)=\mathbb{E}_{r}[\epsilon_k^2]=\frac{\|a\|_2^2}{d}\sum_{i\neq k}\|v_i\|_2^2$,
and therefore:
\begin{equation}
\label{eq:interference_std}
\small
\operatorname{Std}_{r}(\epsilon_k)
=
\frac{\|a\|_2}{\sqrt{d}}
\left(
\sum_{i\neq k}\|v_i\|_2^2
\right)^{1/2}
\end{equation}
Although the full interference vector can retain substantial energy, what matters to subsequent layers is often its projection onto fixed feature directions. Randomized binding disperses interference across the high-dimensional channel space, making it weakly aligned with any particular direction and thereby approximately preserving linear features of the original latent code. At the full-representation level,$\mathbb{E}_r\|\eta_k\|_F^2=\sum_{i\neq k}\|s_i\|_F^2$,
so increasing the number of superposed slots increases total interference energy when code norms remain comparable. Thus, randomized binding does not make interference small in norm; rather, it makes its contribution along any fixed feature direction zero-mean and dimensionally dispersed. This analysis characterizes the binding stage before learned adaptation and does not predict the nonlinear recovery difficulty as $K$ increases. The storage adapter and recovery network are trained to exploit this structure, while Section~\ref{sec:experiments} empirically measures the resulting recovery penalty under more aggressive superposition.

\textbf{The capacity--interference regime: }
The central hypothesis of SLAE is that superposition is useful when the information gained from a wider latent exceeds the cost of recovering it from shared memory. Consider a conventional autoencoder at storage budget $B$, an independently stored wide autoencoder using the $KB$-scalar latent available to SLAE, and SLAE at the original average storage budget $B$. Let $\mathcal{E}_{B}$, $\mathcal{E}_{KB}$, and $\mathcal{E}_{\mathrm{SLAE}}$ denote their reconstruction MSEs. We define the capacity gain from using the wider latent as $G_{\rm cap}=\mathcal{E}_{B}-\mathcal{E}_{KB}$, and the effective superposition--recovery penalty as $P_{\rm sup}=\mathcal{E}_{\mathrm{SLAE}}-\mathcal{E}_{KB}$. The realized matched-storage gain is therefore $\Delta_{\mathrm{SLAE}}=G_{\rm cap}-P_{\rm sup}=\mathcal{E}_{B}-\mathcal{E}_{\mathrm{SLAE}}$. This decomposition makes the operating regime of SLAE intuitive. A wider latent is useful when it recovers information that would otherwise be lost to the narrow bottleneck, but superposition introduces a recovery cost. SLAE improves when the former outweighs the latter. We therefore expect the strongest gains when latent capacity is limiting and superposition remains recoverable, while more aggressive superposition eventually becomes interference limited. Section~\ref{sec:experiments} tests this tradeoff across datasets, storage budgets, and superposition factors.

\section{Experiments}
\label{sec:experiments}

We evaluate SLAE to answer four questions. \textbf{(1)} Does superposed latent storage improve reconstruction quality over a conventional autoencoder at the same average memory budget? \textbf{(2)} Do these gains follow the capacity--interference regime identified in Section~\ref{sec:analysis}? \textbf{(3)} Are the improvements attributable to superposed storage rather than additional model capacity? \textbf{(4)} Does the information preserved by SLAE remain useful beyond reconstruction?

\textbf{Experimental setup: } We evaluate on CIFAR-10/100~\citep{krizhevsky2009learning}, SVHN~\citep{netzer2011reading}, STL-10~\citep{coates2011analysis}, and Tiny ImageNet~\citep{le2015tiny}, spanning different numbers of classes, image resolutions, and reconstruction difficulty. Across datasets, we vary the latent spatial size $H$, storage budget $B$, and superposition factor $K\in\{2,4,8\}$ to cover both strongly bottlenecked and more expressive latent regimes. Reconstruction quality is measured using MSE and SSIM, with MSE used as the primary error metric in analyses. Our primary reconstruction experiments follow a corpus-compression setting: the official evaluation split is treated as the unlabeled target corpus whose representations are to be stored, rather than as a held-out test of autoencoder generalization.

For each configuration, the conventional autoencoder stores a latent in $\mathbb{R}^{C\times H\times H}$ using $B=CH^2$ scalars per example. SLAE instead uses a $K$-times wider latent in $\mathbb{R}^{KC\times H\times H}$ and superposes groups of $K$ examples, giving the same average storage $B$ per example. Examples are randomly grouped. We then compare Plain and SLAE at exactly matched average latent storage. For the global best-under-budget comparison, we search $H\in\{2,3,4\}$ for CIFAR-10, CIFAR-100, and SVHN, and $H\in\{4,5,6\}$ for STL-10 and Tiny ImageNet, with channel widths $C\in\{8,16,32,64,128,256\}$. The storage budget $B$ counts per-example latent values; model parameters and globally reused binding keys are fixed overheads and do not scale with the number of stored examples. All methods use the same precision, so matched scalar count also corresponds to matched storage in bits. SLAE is complementary to precision reduction. Combining moderate superposition with quantized shared-memory storage improves the memory--quality tradeoff at aggressive compression (Appendix~\ref{app:alternative_compression}).

Plain and SLAE use the same underlying encoder--decoder architecture for each dataset and latent geometry. We use \emph{Wide Plain AE} to denote a conventional AE with the same $KC\times H\times H$ latent geometry as SLAE; it requires $K$ times the storage and is used only as a capacity reference and for initialization. SLAE initializes its encoder and decoder from the corresponding Wide Plain autoencoder and then trains the storage adapter and recovery pathway as described in Section~\ref{sec:method}. Each $(K,C,H)$ configuration is trained separately.

\begin{figure}[t]
    \centering
    \includegraphics[width=0.8\textwidth]{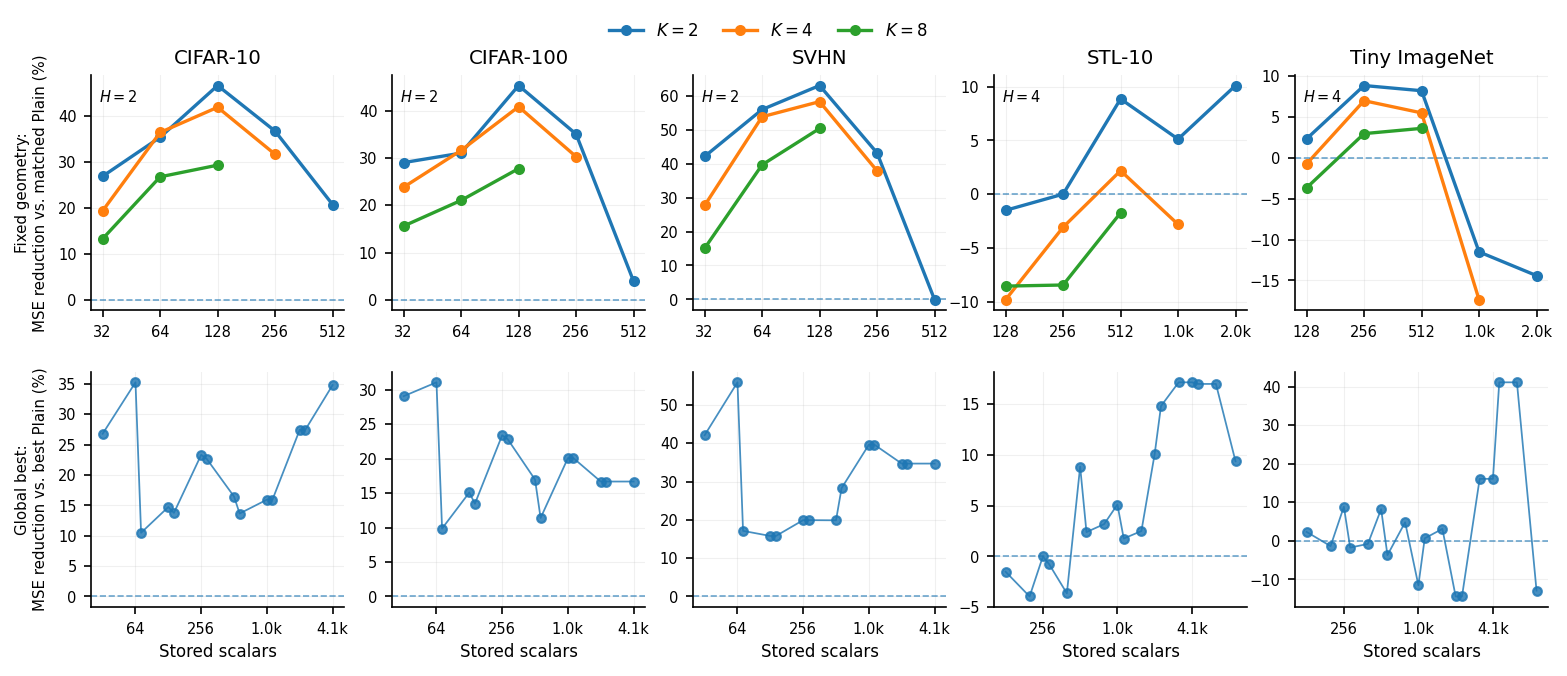}
    \vspace{-5mm}
    \caption{\textbf{Reconstruction improvement at matched latent storage.}
    Relative MSE reduction of SLAE over Plain at the same average number of stored scalars per example. \textbf{Top:} controlled fixed-geometry comparison for $K\in\{2,4,8\}$. \textbf{Bottom:} global best-under-budget comparison, where each method selects its best available latent configuration. Positive values indicate an SLAE improvement. Lines in the bottom row connect independently selected configurations only as a visual guide. Y-axis limits vary across datasets for readability.}
    \label{fig:matched_storage}
    \vspace{-6mm}
\end{figure}

\begin{wraptable}{r}{0.40\columnwidth}
    \centering
    \vspace{-7mm}
    \caption{\textbf{Capped storage.} Best model configuration under $B=512$ cap.}
    \label{tab:global_512}
    \vspace{-0.1em}
    \scriptsize
    \setlength{\tabcolsep}{2.5pt}
    \begin{tabular}{lccc}
        \toprule
        Dataset & Plain & SLAE & Relative MSE Red. \\
        \midrule
        CIFAR-10  & .000833 & \textbf{.000696} & \textbf{16.5\%} \\
        CIFAR-100 & .000894 & \textbf{.000742} & \textbf{17.0\%} \\
        SVHN      & .000050 & \textbf{.000040} & \textbf{19.9\%} \\
        STL-10    & .005248 & \textbf{.004782} & \textbf{8.9\%} \\
        TinyImage & .008245 & \textbf{.007568} & \textbf{8.2\%} \\
        \bottomrule
    \end{tabular}
    \vspace{-1em}
\end{wraptable}
\textbf{Reconstruction at matched storage:} 
We first compare SLAE with conventional autoencoders at the same average latent-storage budget. Figure~\ref{fig:matched_storage} reports the relative MSE reduction of SLAE over Plain. The top row fixes a latent spatial size $H$, isolating the effect of replacing channel reduction with shared storage of wider latents. The bottom row gives the stricter global comparison, where each method selects its best available latent configuration $(H,C)$ under each storage budget. At fixed geometry, SLAE produces large improvements on CIFAR-10, CIFAR-100, and SVHN, exceeding $40$--$60\%$ MSE reduction in several strongly bottlenecked settings. These gains remain substantial under the global comparison, reaching approximately $35\%$, $31\%$, and $56\%$, respectively, with SLAE improving over Plain at all evaluated storage budgets. STL-10 and Tiny ImageNet are more regime-dependent. SLAE improves at $13/18$ and $10/18$ global budget points, respectively, with peak MSE reductions of approximately $17\%$ and $41\%$. At a common budget of $B=512$ scalars per example, SLAE improves reconstruction on all five datasets even when each method selects its best latent geometry, reducing MSE by $8.2$--$19.9\%$ (Table~\ref{tab:global_512}).

\begin{figure}
    \centering
    \vspace{-3mm}
    \includegraphics[width=0.7\linewidth]{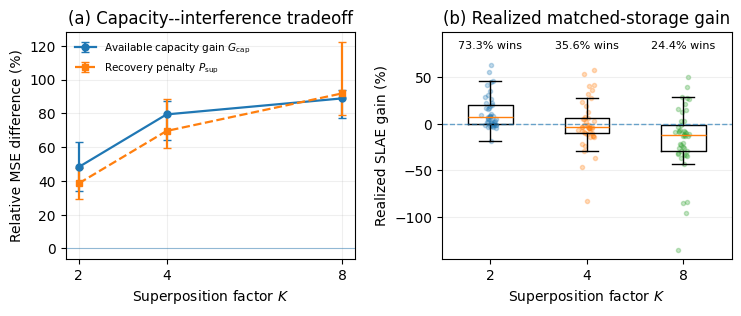}
    \vspace{-4mm}
    \caption{\textbf{Capacity--interference tradeoff as superposition increases.}
    Results use the same 45 $(\mathrm{dataset},H,B)$ configurations for all $K\in\{2,4,8\}$. \textbf{Left:} median capacity gain $G_{\rm cap}$ and recovery penalty $P_{\rm sup}$, with interquartile ranges. \textbf{Right:} distribution of the realized matched-storage gain $\Delta_{\mathrm{SLAE}}=G_{\rm cap}-P_{\rm sup}$; positive values indicate an improvement over Plain.}
    \label{fig:capacity_interference_K}
    \vspace{-4mm}
\end{figure}

\textbf{When does SLAE help: }

The matched-storage results above show that SLAE's advantage is strongly regime-dependent, motivating the capacity--interference analysis of Section~\ref{sec:analysis}. Increasing the superposition factor $K$ preserves progressively wider latent representations, increasing the capacity gain $G_{\rm cap}$ available over a narrow Plain latent. At the same time, recovering each latent from a more heavily superposed memory becomes increasingly difficult, increasing the recovery penalty $P_{\rm sup}$. For each configuration, their difference determines the realized matched-storage gain, $\Delta_{\mathrm{SLAE}}=G_{\rm cap}-P_{\rm sup}=\mathcal{E}_{B}-\mathcal{E}_{\mathrm{SLAE}}$. Figure~\ref{fig:capacity_interference_K} tests this tradeoff on 45 configurations shared across all three values of $K$. For comparisons across datasets, we normalize $G_{\rm cap}$, $P_{\rm sup}$, and $\Delta_{\rm SLAE}$ by $\mathcal{E}_B$ and report them as percentages. The median capacity gain increases from $48.2\%$ at $K=2$ to $79.3\%$ at $K=4$ and $88.8\%$ at $K=8$. However, the median recovery penalty rises from $38.5\%$ to $69.5\%$ and then $91.8\%$. As a result, the median realized SLAE gain shifts from $+7.7\%$ at $K=2$ to $-3.9\%$ at $K=4$ and $-12.3\%$ at $K=8$, while the fraction of configurations in which SLAE outperforms Plain falls from $73.3\%$ to $35.6\%$ and $24.4\%$. Because we evaluate the same configurations for every $K$, we can directly measure the effect of increasing superposition. From $K=2$ to $K=8$, the capacity gain increases in $97.8\%$ of configurations, but the recovery penalty increases in all of them and the realized SLAE gain decreases. Thus, larger $K$ provides more latent capacity, but also introduces more recovery difficulty, making moderate superposition, particularly $K=2$ in our experiments, the most reliable operating regime.

\begin{wrapfigure}{r}{0.46\columnwidth}
    \centering
    \vspace{-4mm}
    \includegraphics[width=\linewidth]{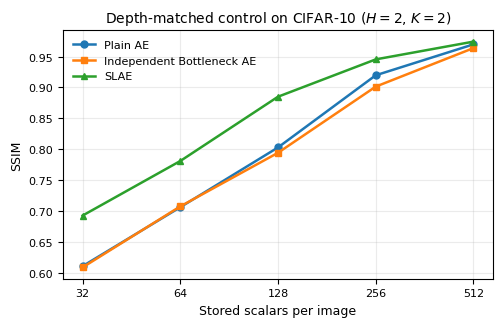}
    \vspace{-9mm}
    \caption{\textbf{Depth-matched control.} CIFAR-10, $H=2$, $K=2$. The Independent Bottleneck AE adds comparable compression machinery but stores examples independently.}
    \label{fig:depth_matched}
    \vspace{-4mm}
\end{wrapfigure}
\textbf{Where do the SLAE gains come from: } To separate the benefit of superposition from that of additional network depth introduced by the storage adapter and recovery module, we compare three storage-matched models. Plain AE directly stores a narrow $C\times H\times H$ latent. \emph{Independent Bottleneck AE} uses the same wider $KC\times H\times H$ latent and additional network blocks as SLAE, but compresses each example independently to a $C\times H\times H$ bottleneck and then expands it again before decoding. This gives it a deeper compression pathway without any cross-example sharing. \emph{SLAE} instead keeps the wider latent and meets the same average storage budget by superposing $K$ examples into shared memory. Figure~\ref{fig:depth_matched} shows that the independent bottleneck does not improve over Plain, while SLAE consistently does across the shared budgets. This isolates superposition as the key difference and indicates that the gain comes from sharing wider latent representations rather than a deeper network.

\begin{wrapfigure}{r}{0.46\columnwidth}
    \centering
    \vspace{-4mm}
    \includegraphics[width=\linewidth]{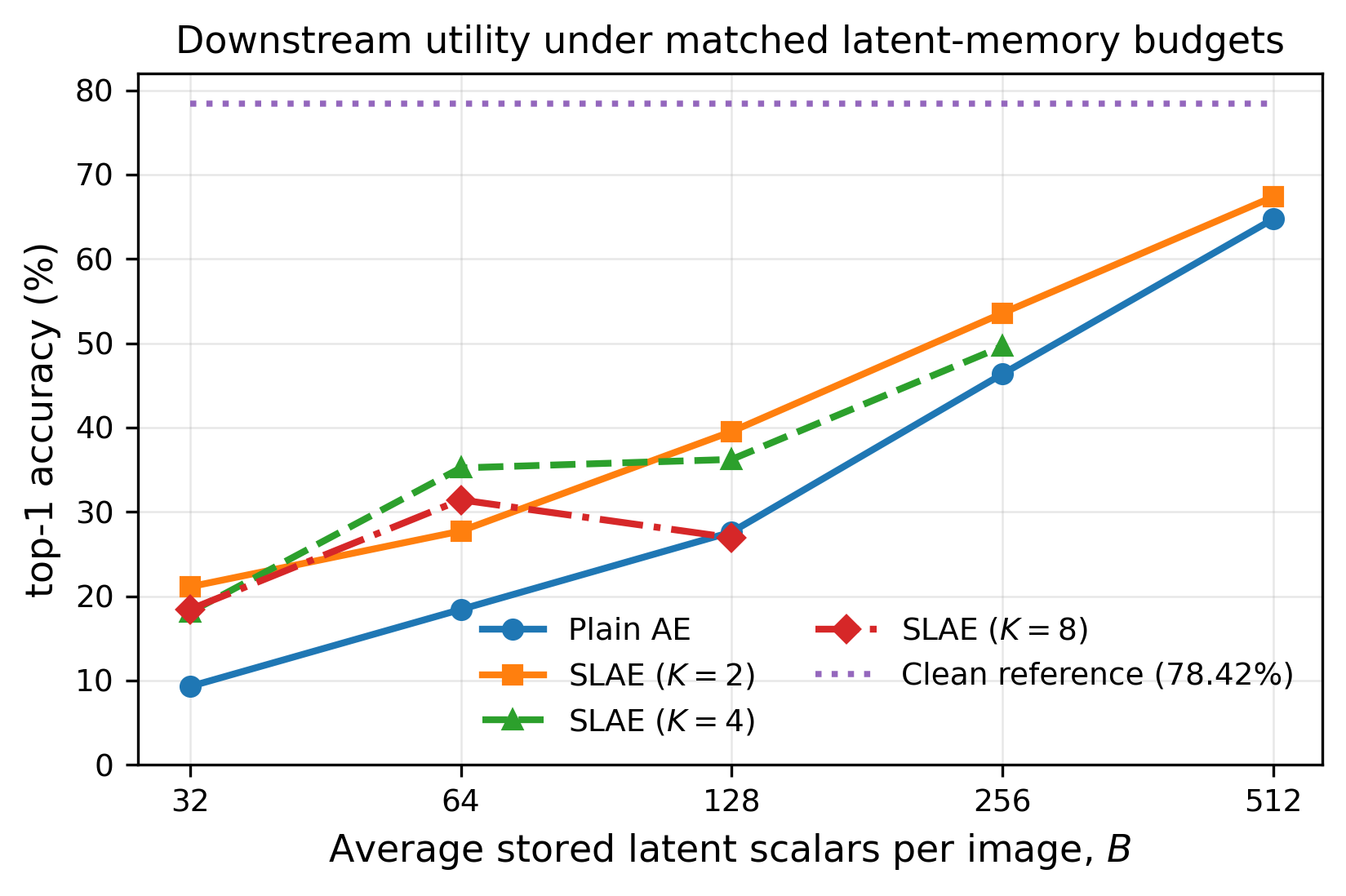}
    \vspace{-7mm}
    \caption{\textbf{Downstream utility at matched storage.}
    CIFAR-100 top-1 accuracy when the same CIFAR-style ResNet-18 is trained on Plain AE or SLAE reconstructions and evaluated on the same clean test set.}
    \label{fig:downstream_utility}
    \vspace{-4mm}
\end{wrapfigure}
\textbf{Downstream utility of stored representations: } We next test whether the information preserved by SLAE remains useful for downstream learning. On CIFAR-100, we compress and reconstruct the training set using Plain AE or SLAE under matched average latent-storage budgets, then
train the same CIFAR-style ResNet-18 from scratch on each reconstructed training set. All classifiers use the same training protocol and are evaluated on the original clean test set; training on the uncompressed data provides an upper reference. Figure~\ref{fig:downstream_utility} shows that SLAE improves over matched-storage Plain AE in 11 of 12 comparisons, including every evaluated $K=2$ and $K=4$ setting. At $B=128$ with $K=2$, top-1 accuracy increases from $27.61\%$ to $39.53\%$, a gain of $11.92$ percentage points. The gains are generally largest under tighter memory budgets and diminish as the Plain representation becomes less capacity-limited. These results show that SLAE's reconstruction improvements translate into substantially greater downstream utility.

\section{Conclusion}

We introduced SLAE, a different way to compress latent representations under fixed memory budget: instead of shrinking each latent independently, SLAE keeps representations wide and lets multiple examples share storage through superposition. This replaces irreversible capacity loss with a structured interference-and-recovery problem. Across five image datasets, SLAE improves the reconstruction--memory tradeoff, reducing MSE by up to $56\%$, with improvements on all five datasets and the strongest gains under tight memory bottlenecks. The preserved information is also useful downstream, improving CIFAR-100 classification by up to $16.79$ percentage points at the same storage budget. More broadly, our results suggest that under tight memory budgets, preserving rich representations and sharing their storage can be better than shrinking each representation in isolation. This principle could have broader implications for memory-constrained representation learning, continual/lifelong learning, latent/vector databases, and other memory-critical systems.

\section*{Acknowledgment}
This work was supported in part by PRISM and CoCoSys, centers in JUMP 2.0, an SRC program sponsored by DARPA (SRC grant number - 2023-JU-3135). This work was also supported by NSF grants \#2003279, \#1911095, \#2112167, \#2052809, \#2112665, \#2120019, \#2211386.


\bibliographystyle{iclr2027_conference}
\bibliography{iclr2027_conference}

@article{hinton2006reducing,
  title={Reducing the dimensionality of data with neural networks},
  author={Hinton, Geoffrey E and Salakhutdinov, Ruslan R},
  journal={science},
  volume={313},
  number={5786},
  pages={504--507},
  year={2006},
  publisher={American Association for the Advancement of Science}
}

@article{bengio2013representation,
  title={Representation learning: A review and new perspectives},
  author={Bengio, Yoshua and Courville, Aaron and Vincent, Pascal},
  journal={IEEE transactions on pattern analysis and machine intelligence},
  volume={35},
  number={8},
  pages={1798--1828},
  year={2013},
  publisher={IEEE}
}

@inproceedings{chen2024compressing,
  title={Compressing latent space via least volume},
  author={Chen, Qiuyi and Fuge, Mark},
  booktitle={International Conference on Learning Representations},
  volume={2024},
  pages={37324--37347},
  year={2024}
}

@article{plate1995holographic,
  title={Holographic reduced representations},
  author={Plate, Tony A},
  journal={IEEE Transactions on Neural networks},
  volume={6},
  number={3},
  pages={623--641},
  year={1995},
  publisher={IEEE}
}

@article{cheung2019superposition,
  title={Superposition of many models into one},
  author={Cheung, Brian and Terekhov, Alexander and Chen, Yubei and Agrawal, Pulkit and Olshausen, Bruno},
  journal={Advances in neural information processing systems},
  volume={32},
  year={2019}
}

@article{menet2023mimonets,
  title={Mimonets: Multiple-input-multiple-output neural networks exploiting computation in superposition},
  author={Menet, Nicolas and Hersche, Michael and Karunaratne, Geethan and Benini, Luca and Sebastian, Abu and Rahimi, Abbas},
  journal={Advances in Neural Information Processing Systems},
  volume={36},
  pages={39553--39565},
  year={2023}
}

@inproceedings{rannen2017encoder,
  title={Encoder based lifelong learning},
  author={Rannen, Amal and Aljundi, Rahaf and Blaschko, Matthew B and Tuytelaars, Tinne},
  booktitle={2017 IEEE international conference on computer vision (ICCV)},
  pages={1329--1337},
  year={2017},
  organization={IEEE}
}

@article{theis2017lossy,
  title={Lossy image compression with compressive autoencoders},
  author={Theis, Lucas and Shi, Wenzhe and Cunningham, Andrew and Husz{\'a}r, Ferenc},
  journal={arXiv preprint arXiv:1703.00395},
  year={2017}
}

@inproceedings{chen2025deep,
  title={Deep compression autoencoder for efficient high-resolution diffusion models},
  author={Chen, Junyu and Cai, Han and Chen, Junsong and Xie, Enze and Yang, Shang and Tang, Haotian and Li, Muyang and Han, Song},
  booktitle={International Conference on Learning Representations},
  volume={2025},
  pages={96539--96560},
  year={2025}
}

@article{wu2025h3ae,
  title={H3ae: High compression, high speed, and high quality autoencoder for video diffusion models},
  author={Wu, Yushu and Li, Yanyu and Skorokhodov, Ivan and Kag, Anil and Menapace, Willi and Girish, Sharath and Siarohin, Aliaksandr and Wang, Yanzhi and Tulyakov, Sergey},
  journal={arXiv preprint arXiv:2504.10567},
  year={2025}
}

@article{kleyko2022vector,
  title={Vector symbolic architectures as a computing framework for emerging hardware},
  author={Kleyko, Denis and Davies, Mike and Frady, Edward Paxon and Kanerva, Pentti and Kent, Spencer J and Olshausen, Bruno A and Osipov, Evgeny and Rabaey, Jan M and Rachkovskij, Dmitri A and Rahimi, Abbas and others},
  journal={Proceedings of the IEEE},
  volume={110},
  number={10},
  pages={1538--1571},
  year={2022},
  publisher={IEEE}
}

@article{psenka2024representation,
  title={Representation learning via manifold flattening and reconstruction},
  author={Psenka, Michael and Pai, Druv and Raman, Vishal and Sastry, Shankar and Ma, Yi},
  journal={Journal of Machine Learning Research},
  volume={25},
  number={132},
  pages={1--47},
  year={2024}
}

@inproceedings{chakraborty2024statistical,
  title={A statistical analysis of wasserstein autoencoders for intrinsically low-dimensional data},
  author={Chakraborty, Saptarshi and Bartlett, Peter},
  booktitle={International Conference on Learning Representations},
  volume={2024},
  pages={11478--11509},
  year={2024}
}

@article{balle2018variational,
  title={Variational image compression with a scale hyperprior},
  author={Ball{\'e}, Johannes and Minnen, David and Singh, Saurabh and Hwang, Sung Jin and Johnston, Nick},
  journal={arXiv preprint arXiv:1802.01436},
  year={2018}
}

@inproceedings{he2022elic,
  title={Elic: Efficient learned image compression with unevenly grouped space-channel contextual adaptive coding},
  author={He, Dailan and Yang, Ziming and Peng, Weikun and Ma, Rui and Qin, Hongwei and Wang, Yan},
  booktitle={2022 IEEE/CVF Conference on Computer Vision and Pattern Recognition (CVPR)},
  pages={5708--5717},
  year={2022},
  organization={IEEE}
}

@article{balle2016end,
  title={End-to-end optimized image compression},
  author={Ball{\'e}, Johannes and Laparra, Valero and Simoncelli, Eero P},
  journal={arXiv preprint arXiv:1611.01704},
  year={2016}
}

@inproceedings{li2024nerfcodec,
  title={Nerfcodec: Neural feature compression meets neural radiance fields for memory-efficient scene representation},
  author={Li, Sicheng and Li, Hao and Liao, Yiyi and Yu, Lu},
  booktitle={2024 IEEE/CVF Conference on Computer Vision and Pattern Recognition (CVPR)},
  pages={21274--21283},
  year={2024},
  organization={IEEE}
}

@article{dong2024get,
  title={Get more with less: Synthesizing recurrence with kv cache compression for efficient llm inference},
  author={Dong, Harry and Yang, Xinyu and Zhang, Zhenyu and Wang, Zhangyang and Chi, Yuejie and Chen, Beidi},
  journal={arXiv preprint arXiv:2402.09398},
  year={2024}
}

@article{kanerva2009hyperdimensional,
  title={Hyperdimensional computing: An introduction to computing in distributed representation with high-dimensional random vectors},
  author={Kanerva, Pentti},
  journal={Cognitive computation},
  volume={1},
  number={2},
  pages={139--159},
  year={2009},
  publisher={Springer}
}

@article{bereska2025superposition,
  title={Superposition as lossy compression: Measure with sparse autoencoders and connect to adversarial vulnerability},
  author={Bereska, Leonard and Tzifa-Kratira, Zoe and Samavi, Reza and Gavves, Efstratios},
  journal={arXiv preprint arXiv:2512.13568},
  year={2025}
}

@article{murahari2022datamux,
  title={Datamux: Data multiplexing for neural networks},
  author={Murahari, Vishvak and Jimenez, Carlos and Yang, Runzhe and Narasimhan, Karthik},
  journal={Advances in Neural Information Processing Systems},
  volume={35},
  pages={17515--17527},
  year={2022}
}

@article{bourlard1988auto,
  title={Auto-association by multilayer perceptrons and singular value decomposition},
  author={Bourlard, Herv{\'e} and Kamp, Yves},
  journal={Biological cybernetics},
  volume={59},
  number={4},
  pages={291--294},
  year={1988},
  publisher={Springer}
}

@article{baldi1989neural,
  title={Neural networks and principal component analysis: Learning from examples without local minima},
  author={Baldi, Pierre and Hornik, Kurt},
  journal={Neural networks},
  volume={2},
  number={1},
  pages={53--58},
  year={1989},
  publisher={Elsevier}
}

@inproceedings{cosentino2022deep,
  title={Deep autoencoders: From understanding to generalization guarantees},
  author={Cosentino, Romain and Balestriero, Randall and Baranuik, Richard and Aazhang, Behnaam},
  booktitle={Mathematical and Scientific Machine Learning},
  pages={197--222},
  year={2022},
  organization={PMLR}
}

@inproceedings{masci2011stacked,
  title={Stacked convolutional auto-encoders for hierarchical feature extraction},
  author={Masci, Jonathan and Meier, Ueli and Cire{\c{s}}an, Dan and Schmidhuber, J{\"u}rgen},
  booktitle={International conference on artificial neural networks},
  pages={52--59},
  year={2011},
  organization={Springer}
}

@inproceedings{he2016deep,
  title={Deep residual learning for image recognition},
  author={He, Kaiming and Zhang, Xiangyu and Ren, Shaoqing and Sun, Jian},
  booktitle={Proceedings of the IEEE conference on computer vision and pattern recognition},
  pages={770--778},
  year={2016}
}

@inproceedings{ailon2006approximate,
  title={Approximate nearest neighbors and the fast Johnson-Lindenstrauss transform},
  author={Ailon, Nir and Chazelle, Bernard},
  booktitle={Proceedings of the thirty-eighth annual ACM symposium on Theory of computing},
  pages={557--563},
  year={2006}
}

@inproceedings{perez2018film,
  title={Film: Visual reasoning with a general conditioning layer},
  author={Perez, Ethan and Strub, Florian and De Vries, Harm and Dumoulin, Vincent and Courville, Aaron},
  booktitle={Proceedings of the AAAI conference on artificial intelligence},
  volume={32},
  number={1},
  year={2018}
}

@article{wang2004image,
  title={Image quality assessment: from error visibility to structural similarity},
  author={Wang, Zhou and Bovik, Alan C and Sheikh, Hamid R and Simoncelli, Eero P},
  journal={IEEE transactions on image processing},
  volume={13},
  number={4},
  pages={600--612},
  year={2004},
  publisher={IEEE}
}

@article{cogswell2015reducing,
  title={Reducing overfitting in deep networks by decorrelating representations},
  author={Cogswell, Michael and Ahmed, Faruk and Girshick, Ross and Zitnick, Larry and Batra, Dhruv},
  journal={arXiv preprint arXiv:1511.06068},
  year={2015}
}

@inproceedings{zbontar2021barlow,
  title={Barlow twins: Self-supervised learning via redundancy reduction},
  author={Zbontar, Jure and Jing, Li and Misra, Ishan and LeCun, Yann and Deny, St{\'e}phane},
  booktitle={International conference on machine learning},
  pages={12310--12320},
  year={2021},
  organization={PMLR}
}

@article{krizhevsky2009learning,
  title={Learning multiple layers of features from tiny images},
  author={Krizhevsky, Alex and Hinton, Geoffrey and others},
  year={2009},
  publisher={Toronto, ON, Canada}
}

@inproceedings{netzer2011reading,
  title={Reading digits in natural images with unsupervised feature learning},
  author={Netzer, Yuval and Wang, Tao and Coates, Adam and Bissacco, Alessandro and Wu, Baolin and Ng, Andrew Y and others},
  booktitle={NIPS workshop on deep learning and unsupervised feature learning},
  volume={2011},
  number={2},
  pages={4},
  year={2011},
  organization={Granada}
}

@inproceedings{coates2011analysis,
  title={An analysis of single-layer networks in unsupervised feature learning},
  author={Coates, Adam and Ng, Andrew and Lee, Honglak},
  booktitle={Proceedings of the fourteenth international conference on artificial intelligence and statistics},
  pages={215--223},
  year={2011},
  organization={JMLR Workshop and Conference Proceedings}
}

@article{le2015tiny,
  title={Tiny imagenet visual recognition challenge},
  author={Le, Yann and Yang, Xuan and others},
  journal={CS 231N},
  volume={7},
  number={7},
  pages={3},
  year={2015}
}

@article{allerbo2021non,
  title={Non-linear, sparse dimensionality reduction via path lasso penalized autoencoders},
  author={Allerbo, Oskar and J{\"o}rnsten, Rebecka},
  journal={Journal of Machine Learning Research},
  volume={22},
  number={283},
  pages={1--28},
  year={2021}
}

@article{wei2022qdrop,
  title={Qdrop: Randomly dropping quantization for extremely low-bit post-training quantization},
  author={Wei, Xiuying and Gong, Ruihao and Li, Yuhang and Liu, Xianglong and Yu, Fengwei},
  journal={arXiv preprint arXiv:2203.05740},
  year={2022}
}

@inproceedings{liu2023pd,
  title={Pd-quant: Post-training quantization based on prediction difference metric},
  author={Liu, Jiawei and Niu, Lin and Yuan, Zhihang and Yang, Dawei and Wang, Xinggang and Liu, Wenyu},
  booktitle={2023 IEEE/CVF Conference on Computer Vision and Pattern Recognition (CVPR)},
  pages={24427--24437},
  year={2023},
  organization={IEEE}
}

@inproceedings{liu2023learned,
  title={Learned image compression with mixed transformer-cnn architectures},
  author={Liu, Jinming and Sun, Heming and Katto, Jiro},
  booktitle={2023 IEEE/CVF conference on computer vision and pattern recognition (CVPR)},
  pages={14388--14397},
  year={2023},
  organization={IEEE}
}

\newpage
\appendix

The appendix here provides additional details for the submission titled: “Superposed Latent Autoencoder”. The appendix is organized as follows:

{\setstretch{2}
\begin{enumerate}[label=\Alph*.]
    \large
    \item \hyperref[app:notations]{\textbf{List of Notations}}
    \item \hyperref[app:qualitative]{\textbf{Qualitative Reconstructions}}
    \item \hyperref[app:component_study]{\textbf{Component Study}}
    \item \hyperref[app:param_overhead]{\textbf{Parameter Overhead}}
    \item \hyperref[app:multiseed]{\textbf{Multi-Seed Stability}}
    \item \hyperref[app:implementation_detail]{\textbf{Implementation and Training Details}}
    \item \hyperref[app:downstream_classification]{\textbf{Downstream Classification Details}}    
    \item \hyperref[app:protocol_robustness]{\textbf{Robustness to Checkpoint-Selection Protocol}}
    \item \hyperref[app:full_reconstruction]{\textbf{Full Reconstruction Results}}
    \item \hyperref[app:capacity_interference]{\textbf{Additional Capacity--Interference Results}}
    \item \hyperref[app:binding_derivation]{\textbf{Derivation of Randomized-Binding Interference}}
    \item \hyperref[app:matched_quality]{\textbf{Memory Savings at Matched Quality}}
    \item \hyperref[app:alternative_compression]{\textbf{Comparison with Alternative Latent Compression Strategies}}
    \item \hyperref[app:limitations]{\textbf{Limitations}}
    \item \hyperref[app:discussion]{\textbf{Discussion}}
\end{enumerate}
}

\newpage

\appendix

\section{List of Notations}
\label{app:notations}

We hereby provide a list of notations used in this paper:

\begin{table*}[h]
\caption{List of notations.}
\begin{center}
\small
\setlength{\tabcolsep}{6pt}
\begin{tabular}{p{10em} || p{32em}} 
 \toprule
 \toprule
 Symbol & Meaning \\
 \midrule

 \centering \(x_i\) & Input example assigned to superposition slot \(i\). \\

 \centering \(E, D\) & Encoder and decoder. \\

 \centering \(z_i = E(x_i)\) & Wide decoder-facing latent representation of input \(x_i\). \\

 \centering \(S, R\) & Storage adapter and recovery network. \\

 \centering \(s_i = S(z_i)\) & Storage-friendly latent code. \\

 \centering \(\hat{s}_i\) & Demixed storage code obtained after inverse binding. \\

 \centering \(\hat{z}_i = R(\hat{s}_i;i)\) & Recovered decoder-facing latent representation. \\

 \centering \(\hat{x}_i = D(\hat{z}_i)\) & Reconstructed input. \\

 \centering \(K\) & Number of examples superposed into one shared memory tensor. \\

 \centering \(C\) & Channel dimension of the matched-storage Plain AE latent. \\

 \centering \(H\) & Spatial latent size; latent tensors have spatial resolution \(H\times H\). \\

 \centering \(B=CH^2\) & Average number of stored latent scalars per example. \\

 \centering \(d=KC\) & Channel dimension of the SLAE storage code at one spatial location. \\

 \centering \(r_i\) & Randomized binding key assigned to slot \(i\). \\

 \centering \(P_i\) & Random channel-permutation matrix associated with slot \(i\). \\

 \centering \(D_i\) & Diagonal Rademacher sign matrix associated with slot \(i\). \\

 \centering \(\mathcal{H}\) & Normalized Walsh--Hadamard transform. \\

 \centering \(B_{r_i}\) & Orthogonal binding operator,
 \(B_{r_i}=\mathcal{H}D_iP_i\). \\

 \centering \(b_i=B_{r_i}(s_i)\) & Bound storage representation for slot \(i\). \\

 \centering \(m\) & Shared superposed memory tensor. \\

 \centering \(\eta_k\) & Aggregate cross-slot interference after unbinding slot \(k\). \\

 \centering \(v_i\) & Channel vector of storage code \(s_i\) at a fixed spatial location. \\

 \centering \(a\) & Fixed linear probe or feature direction in channel space. \\

 \centering \(\epsilon_k\) & Scalar interference observed along direction \(a\). \\

 \centering \(E_B\) & Reconstruction MSE of the matched-storage Plain AE at budget \(B\). \\

 \centering \(E_{KB}\) & Reconstruction MSE of the Wide Plain AE using budget \(KB\). \\

 \centering \(E_{\mathrm{SLAE}}\) & Reconstruction MSE of SLAE at average storage budget \(B\). \\

 \centering \(G_{\mathrm{cap}}=E_B-E_{KB}\) & Capacity gain obtained from using the wider latent representation. \\

 \centering \(P_{\mathrm{sup}}=E_{\mathrm{SLAE}}-E_{KB}\) & Effective superposition--recovery penalty. \\

 \centering \(\Delta_{\mathrm{SLAE}}=E_B-E_{\mathrm{SLAE}}\) & Realized matched-storage improvement of SLAE over Plain AE. \\

 \bottomrule
 \bottomrule
\end{tabular}
\label{notation}
\end{center}
\end{table*}

\section{Qualitative Reconstructions}
\label{app:qualitative}
\begin{figure}[H]
    \centering
    \includegraphics[width=\textwidth]{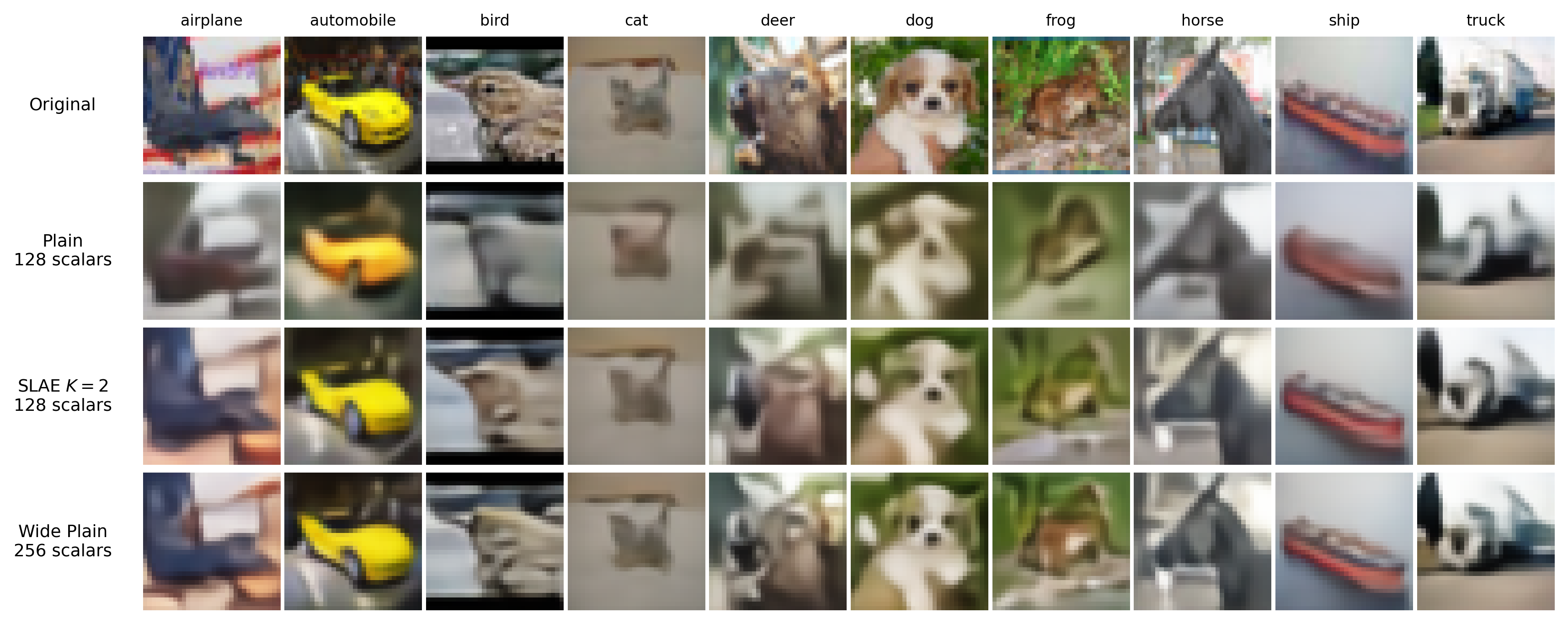}
    \caption{\textbf{Qualitative reconstructions on CIFAR-10.}
    Reconstructions at $H=2$ and $K=2$. Plain AE and SLAE both use an average
    storage budget of $128$ scalars per example. Wide Plain uses the same wide
    latent geometry as SLAE but stores each latent independently and therefore
    requires $256$ scalars per example.}
    \label{fig:qualitative_cifar10}
\end{figure}

\begin{figure}[H]
    \centering
    \includegraphics[width=\textwidth]{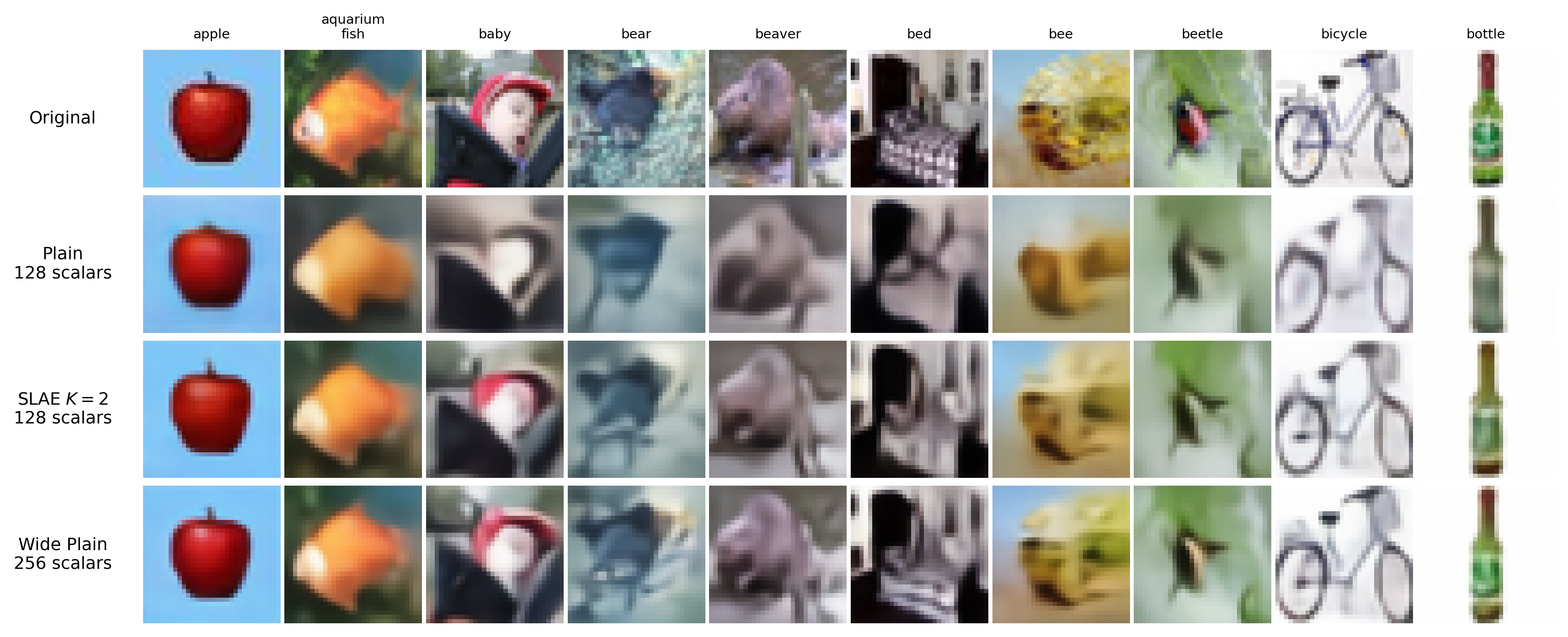}
    \caption{\textbf{Qualitative reconstructions on CIFAR-100.}
    Reconstructions at $H=2$ and $K=2$. Plain AE and SLAE are matched at
    $128$ stored scalars per example, while Wide Plain uses the corresponding
    wider latent independently and requires $256$ scalars per example.}
    \label{fig:qualitative_cifar100}
\end{figure}

Figures~\ref{fig:qualitative_cifar10} and \ref{fig:qualitative_cifar100} provide qualitative comparisons between Plain AE, SLAE, and the corresponding Wide Plain AE under a tight storage regime. Plain AE and SLAE are matched at an average storage budget of $B=128$ scalars per example. For $K=2$, SLAE uses a latent that is twice as wide and stores pairs of examples through superposition, so its average storage remains $128$ scalars per example. Wide Plain uses the same wide latent geometry as SLAE but stores each latent independently, requiring $256$ scalars per example, and therefore serves only as a capacity reference.

Across both datasets, the qualitative reconstructions are consistent with the quantitative results in Section~\ref{sec:experiments}. Relative to the storage-matched Plain AE, SLAE generally retains more of the object structure, appearance, and local detail present in the original image. Its reconstructions also often approach those produced by Wide Plain, despite using only half of Wide Plain's per-example storage. This illustrates the central behavior targeted by SLAE: shared storage can preserve part of the representational advantage of wider latent without increasing the average memory budget.

\section{Component Study}
\label{app:component_study}

We study the main components of SLAE to examine their contributions to reconstruction under superposed storage. We use CIFAR-10 with $H=2$, $K=2$, and a latent channel dimension $C=64$, corresponding to an average storage budget of $B=128$ scalars per example. All variants are evaluated under the same storage setting and training protocol.

\begin{table}[H]
\centering
\caption{\textbf{Component study of SLAE.}
Ablations are evaluated under the same setting as Full SLAE.
$\Delta$MSE reports the relative change in reconstruction MSE compared with
the full model, lower MSE and higher SSIM/PSNR are better.}
\label{tab:component_ablation}
\small
\setlength{\tabcolsep}{5pt}
\begin{tabular}{lcccc}
\toprule
Model & SSIM $\uparrow$ & PSNR $\uparrow$ & MSE $\downarrow$ & $\Delta$MSE \\
\midrule
Full SLAE                   & 0.8845 & 24.11 & 0.003886 & -- \\
w/o Storage Adapter         & 0.8750 & 23.69 & 0.004281 & $+10.2\%$ \\
w/o Recovery                & 0.8820 & 24.00 & 0.003986 & $+2.6\%$ \\
w/o Binding                 & 0.6287 & 14.77 & 0.033397 & $+759.5\%$ \\
w/o Slot Conditioning       & 0.8847 & 24.11 & 0.003885 & $-0.02\%$ \\
\bottomrule
\end{tabular}
\end{table}

Table~\ref{tab:component_ablation} shows that randomized binding is essential for effective shared storage. Removing binding causes reconstruction quality to collapse, increasing MSE by more than $7\times$ relative to Full SLAE and reducing SSIM from $0.8845$ to $0.6287$. This confirms that simply adding multiple latent representations into shared memory is insufficient, the slot-specific randomized transforms are necessary to structure cross-slot interference.

The learned storage adapter also makes a meaningful contribution. Removing it increases reconstruction MSE by $10.2\%$, supporting the design choice of separating the decoder-facing latent from the representation used for superposition. Removing the learned recovery module produces a smaller but consistent degradation of $2.6\%$ in MSE, indicating that learned post-demixing adaptation further improves reconstruction beyond inverse binding alone.

In contrast, removing slot conditioning has negligible effect in this setting. The shared recovery network therefore appears capable of handling the slot-specific interference patterns without requiring an explicit slot identifier. We retain slot conditioning in the full architecture because it provides a lightweight mechanism for slot-specific adaptation, but the ablation indicates that the main gains of SLAE do not depend on this component.

\section{Parameter Overhead}
\label{app:param_overhead}

SLAE augments the underlying encoder--decoder with a learned storage adapter $S$ and recovery network $R$. These modules introduce additional learned parameters beyond the encoder and decoder. To quantify this cost, we count parameters using the exact architectures and hyperparameters used in the main experimental grid. For each SLAE configuration, we define the SLAE-specific parameter overhead as
\begin{equation}
\mathrm{Overhead}
=
100\,
\frac{|S|+|R|}
{|E|+|D|}    
\end{equation}
where $E$ and $D$ are the encoder and decoder of the corresponding Wide Plain AE with the same wide latent geometry. The randomized binding keys are fixed global metadata rather than learned parameters and are not included in this quantity.

\begin{table}[H]
\centering
\caption{\textbf{SLAE-specific learned parameter overhead.}
Entries report $(|S|+|R|)/(|E|+|D|)$ as a percentage for the exact
architectures used in the main experiments. The overhead is small at moderate
latent widths and increases at very large $C$ because the current recovery
network scales with channel dimension.}
\label{tab:param_overhead}
\small
\setlength{\tabcolsep}{5pt}
\begin{tabular}{lccccc}
\toprule
Architecture / latent regime
& $C=16$ & $C=32$ & $C=64$ & $C=128$ & $C=256$ \\
\midrule
CIFAR/SVHN, $H=2$
& 0.8\% & 1.6\% & 4.3\% & 14.1\% & 51.1\% \\
CIFAR/SVHN, $H=3,4$
& 1.2\% & 2.4\% & 6.7\% & 22.0\% & 79.6\% \\
STL-10, $H=4,5,6$
& 1.0\% & 2.2\% & 6.1\% & 19.9\% & 71.4\% \\
Tiny ImageNet, $H=4$
& 1.0\% & 2.2\% & 6.1\% & 19.9\% & 71.4\% \\
Tiny ImageNet, $H=5,6$
& 1.6\% & 3.5\% & 9.5\% & 31.0\% & 111.1\% \\
\bottomrule
\end{tabular}
\end{table}

Across the full experimental grid, the median SLAE-specific learned overhead is approximately $6.7\%$. More importantly, the overhead is modest in the capacity-limited regimes where SLAE is most useful. For example, the CIFAR $H=2$, $K=2$, $B=128$ setting uses a wide latent with $C=64$ and incurs only $4.3\%$ additional learned parameters, while the Tiny ImageNet $H=4$, $K=2$, $B=512$ setting incurs $6.1\%$. These are representative of the tight-memory regimes in which preserving additional latent capacity yields the largest reconstruction benefit.

The overhead increases substantially at very large channel widths. In particular, the $C=256$ configurations incur considerably higher parameter cost because the convolutional and token-mixing layers in the current recovery network grow with latent channel dimension.

Finally, the reconstruction gains cannot be attributed solely to this additional model capacity. The depth-matched Independent Bottleneck AE control (Section~\ref{sec:experiments}) adds comparable compression machinery while storing examples independently, but does not reproduce the gains of SLAE. Together, these results indicate that SLAE achieves its strongest memory--reconstruction improvements in regimes where its additional learned parameter overhead is relatively small, while highlighting parameter-efficient recovery at very large latent widths as a direction for future work.

The component study in Appendix~\ref{app:component_study} further suggests that useful superposition does not require the full learned overhead of the default architecture. Removing the storage adapter increases MSE by 10.2\%, while removing the recovery network increases it by only 2.6\%, and both variants retain most of the benefit of full SLAE. Since the binding--superposition--unbinding operations themselves are parameter-free, the ablations suggest that SLAE's core memory advantage does not depend on substantial additional learned capacity, leaving substantial room for lighter storage and recovery modules with improved parameter efficiency.

\section{Multi-Seed Stability}
\label{app:multiseed}

We evaluate the stability of SLAE across five random seeds on CIFAR-10 using
$H=2$ and $K=2$. The matched Plain AE uses $C=32$, while SLAE uses the
corresponding wider $C=64$ latent and superposes pairs of examples, giving both
methods the same average storage budget of $B=128$ scalars per example.
We additionally report Wide Plain with $C=64$, which stores each wide latent
independently and therefore requires $256$ scalars per example.

\begin{table}[H]
\centering
\caption{\textbf{Multi-seed stability on CIFAR-10.}
Mean $\pm$ standard deviation over five seeds. Plain AE and SLAE are matched
at $B=128$ scalars per example. Wide Plain uses $256$ scalars per example.
Lower MSE and higher SSIM/PSNR are better.}
\label{tab:multiseed}
\small
\setlength{\tabcolsep}{7pt}
\begin{tabular}{lccc}
\toprule
Method & SSIM $\uparrow$ & MSE $\downarrow$ & PSNR $\uparrow$ \\
\midrule
Plain AE
& $0.7676 \pm 0.0014$
& $0.00879 \pm 0.00009$
& $20.56 \pm 0.05$ \\
SLAE
& $\mathbf{0.8817 \pm 0.0007}$
& $\mathbf{0.00402 \pm 0.00003}$
& $\mathbf{23.96 \pm 0.03}$ \\
Wide Plain
& $0.9140 \pm 0.0008$
& $0.00280 \pm 0.00005$
& $25.53 \pm 0.07$ \\
\bottomrule
\end{tabular}
\end{table}

SLAE consistently outperforms the storage-matched Plain AE across all five
seeds and on all three reconstruction metrics. The paired MSE reduction is
$54.30\%\pm0.43\%$, with individual-seed reductions ranging only from
$53.76\%$ to $54.92\%$. SLAE also improves SSIM by
$0.1141\pm0.0014$ and PSNR by $3.40\pm0.04$ dB. In SSIM, it recovers
$77.9\%\pm0.2\%$ of the gap between the matched Plain AE and the
$2\times$-storage Wide Plain reference. The small variance across seeds and
the consistent $5/5$ wins indicate that the matched-storage advantage of SLAE
is not sensitive to a particular random seed.

\section{Implementation and Training Details}
\label{app:implementation_detail}

\subsection{Network Architecture}

\textbf{Encoder and decoder: } Plain AE, Wide Plain AE, and SLAE use the same underlying convolutional encoder--decoder architecture for a given dataset and latent geometry. The encoder consists of residual convolutional blocks with GroupNorm and SiLU activations, with spatial downsampling performed by stride-$2$ $3\times3$ convolutions. Channel width increases through the encoder tower from 64 to 128 and then 256 channels before a final projection to the requested latent channel dimension. Adaptive average pooling is used when necessary to produce the exact target latent spatial size $H\times H$. The decoder mirrors this construction using residual convolutional blocks and interpolation followed by convolution for spatial upsampling. Thus, changing the latent geometry changes the bottleneck size without changing the basic encoder--decoder design.

\textbf{Storage adapter: } The storage adapter $S$ preserves the spatial and channel dimensions of the encoder latent. In the main experiments, $S$ contains one residual block with hidden-width multiplier $2$. The residual branch consists of GroupNorm and SiLU followed by a $1\times1$ channel expansion and a $3\times3$ projection back to the latent width. The final projection is zero-initialized, so $S$ begins as an exact identity map and learns only the transformation needed to make the latent more suitable for superposition and recovery.

\textbf{Recovery network: } The recovery network $R$ combines residual convolutional processing with lightweight mixing across the $H^2$ spatial latent positions. For CIFAR-10, CIFAR-100, and SVHN, $R$ uses two convolutional recovery blocks and two token-mixing blocks. For STL-10 and Tiny ImageNet, we use three convolutional recovery blocks and four token-mixing blocks. The hidden-width multiplier is $2$ in all cases. The recovery network is initialized as an identity map and is conditioned on the retrieval slot $k$ through a learned slot embedding and FiLM-style modulation, which produces feature-wise scale and shift parameters for slot-dependent adaptation. The same recovery network is shared across all slots. A summary of the architecture used for different datasets is shown in Table~\ref{tab:implementation_architecture}.

\begin{table}[H]
\centering
\caption{\textbf{SLAE architecture settings used in the main experiments.}
The storage adapter configuration is shared across all datasets, while a slightly deeper recovery network is used for the higher-resolution datasets.}
\label{tab:implementation_architecture}
\small
\setlength{\tabcolsep}{5pt}
\begin{tabular}{lccccc}
\toprule
Dataset & Resolution & $H$ & $S$ blocks & $R$ conv. & $R$ token \\
\midrule
CIFAR-10    & $32\times32$ & $2,3,4$ & 1 & 2 & 2 \\
CIFAR-100   & $32\times32$ & $2,3,4$ & 1 & 2 & 2 \\
SVHN        & $32\times32$ & $2,3,4$ & 1 & 2 & 2 \\
STL-10      & $96\times96$ & $4,5,6$ & 1 & 3 & 4 \\
Tiny ImageNet & $64\times64$ & $4,5,6$ & 1 & 3 & 4 \\
\bottomrule
\end{tabular}
\end{table}

\textbf{Binding implementation: } Binding is performed independently at every spatial latent location over the channel dimension. Each slot is assigned a fixed random channel permutation and an independent Rademacher sign vector. After permutation and sign flipping, the channel vector is mixed using a normalized fast Walsh--Hadamard transform (FWHT). The inverse operation applies the corresponding inverse transform, sign pattern, and permutation. The keys are generated once from a fixed key seed and reused globally across all superposition groups. Because the binding acts only over channels, it preserves the $H\times H$ spatial organization of the latent representation. All channel dimensions used for binding are powers of two, allowing the Hadamard transform to be implemented efficiently without explicitly storing a dense transform matrix.

\subsection{Hyperparameters and Training}

\textbf{Training procedure: } For each SLAE configuration, we first train the corresponding Wide Plain AE with the same wide latent geometry $KC\times H\times H$. The resulting encoder and decoder parameters initialize $E$ and $D$ in SLAE, while the storage adapter $S$ and recovery network $R$ use their identity initialization described above. SLAE is then optimized in two stages. During an initial clean-path warmup, examples are passed through $E\rightarrow S\rightarrow R\rightarrow D$ without superposition, allowing the storage and recovery transformations to adapt while preserving the pretrained reconstruction representation. The remaining epochs use the full $K$-way superposed pathway and optimize $E$, $D$, $S$, and $R$ jointly. The clean pathway remains active as an auxiliary reconstruction anchor during this second stage.

We use AdamW with weight decay $10^{-4}$ and gradient-norm clipping at $1.0$. Learning rates are assigned separately to $E$, $D$, $S$, and $R$ and remain fixed during training; no learning-rate scheduler is used. Table~\ref{tab:training_hyperparameters} summarizes the dataset-specific settings. For CIFAR-10, CIFAR-100, and SVHN, training runs for 200 epochs, of which the first 20 epochs constitute the clean-path warmup. STL-10 and Tiny ImageNet use 120 epochs with a 6-epoch clean warmup.

\begin{table}[H]
\centering
\caption{\textbf{Main SLAE training hyperparameters.}
The group batch size counts superposition groups rather than individual images. Entries $a/b/c$ give the group batch sizes for $K=2/4/8$, respectively.}
\label{tab:training_hyperparameters}
\small
\setlength{\tabcolsep}{4.5pt}
\begin{tabular}{lcccccccc}
\toprule
Dataset & Epochs & Clean warmup & Batch & Group batch & $\mathrm{lr}_E$ & $\mathrm{lr}_D$ & $\mathrm{lr}_S$ & $\mathrm{lr}_R$ \\
\midrule
CIFAR-10/100 & 200 & 20 & 256 & 128/128/128 &
$1{\times}10^{-5}$ & $1{\times}10^{-5}$ &
$1{\times}10^{-4}$ & $1{\times}10^{-4}$ \\
SVHN & 200 & 20 & 256 & 128/128/128 &
$1{\times}10^{-5}$ & $1{\times}10^{-5}$ &
$1{\times}10^{-4}$ & $1{\times}10^{-4}$ \\
STL-10 & 120 & 6 & 128 & 96/64/32 &
$1{\times}10^{-5}$ & $3{\times}10^{-5}$ &
$2{\times}10^{-4}$ & $3{\times}10^{-4}$ \\
Tiny ImageNet & 120 & 6 & 128 & 96/64/32 &
$1{\times}10^{-5}$ & $3{\times}10^{-5}$ &
$2{\times}10^{-4}$ & $3{\times}10^{-4}$ \\
\bottomrule
\end{tabular}
\end{table}

\textbf{Training objective: } For all datasets, the image reconstruction loss is
\begin{equation}
\ell_{\mathrm{rec}}(\hat{x},x)
=
\mathrm{MAE}(\hat{x},x)
+
\mathrm{MSE}(\hat{x},x)
+
0.5\bigl(1-\mathrm{SSIM}(\hat{x},x)\bigr)
\end{equation}
where
\begin{equation}
\mathrm{MAE}(\hat{x},x)
=
\frac{1}{P}\sum_{p=1}^{P}|\hat{x}_p-x_p|,
\qquad
\mathrm{MSE}(\hat{x},x)
=
\frac{1}{P}\sum_{p=1}^{P}(\hat{x}_p-x_p)^2
\end{equation}
with the average taken over all pixels, channels, and examples in the batch. SSIM is computed with an $11\times11$ Gaussian window with standard deviation $1.5$ and image range $[0,1]$.

During the superposed training stage, the complete objective is
\begin{equation}
\mathcal{L}
=
\lambda_x\mathcal{L}_{\mathrm{sup}}
+
\lambda_z\mathcal{L}_{\mathrm{latent}}
+
\lambda_{\mathrm{clean}}\mathcal{L}_{\mathrm{clean}}
+
\lambda_{\mathrm{decor}}\mathcal{L}_{\mathrm{decor}}
\end{equation}
The superposed reconstruction term is
\begin{equation}
\mathcal{L}_{\mathrm{sup}}
=
\frac{1}{K}\sum_{i=1}^{K}
\ell_{\mathrm{rec}}(\hat{x}_i,x_i)
\end{equation}
where $\hat{x}_i=D(\hat{z}_i)$ is reconstructed after superposition, demixing, and recovery. The latent-recovery term directly encourages the recovered latent to match the corresponding encoder latent,
\begin{equation}
\mathcal{L}_{\mathrm{latent}}
=
\frac{1}{K}\sum_{i=1}^{K}
\mathrm{MSE}(\hat{z}_i,z_i)
\end{equation}
The clean-path term applies the same reconstruction objective without superposition,
\begin{equation}
\mathcal{L}_{\mathrm{clean}}
=
\ell_{\mathrm{rec}}(\hat{x}^{\mathrm{clean}},x^{\mathrm{clean}})
\end{equation}
where the clean pathway is
$E\rightarrow S\rightarrow R\rightarrow D$.

For the decorrelation term, the storage codes are reshaped so that every example, slot, and spatial position forms one observation over the latent channels. Let
$Y\in\mathbb{R}^{N\times d}$ denote this matrix, where $d$ is the storage-code channel dimension. Each channel is standardized as
\begin{equation}
X_{nc}
=
\frac{Y_{nc}-\mu_c}{\sigma_c+\epsilon},
\qquad
\epsilon=10^{-6}
\end{equation}
and the resulting channel correlation matrix is
\begin{equation}
C=\frac{X^\top X}{N-1}
\end{equation}
We then penalize squared off-diagonal correlations,
\begin{equation}
\mathcal{L}_{\mathrm{decor}}
=
\frac{1}{d^2}
\sum_{p\neq q} C_{pq}^2
\end{equation}

The loss weights used in the main experiments are summarized below:
\begin{table}[H]
\centering
\caption{\textbf{Loss hyperparameters used in the main experiments.}}
\label{tab:loss_hyperparameters}
\small
\setlength{\tabcolsep}{5pt}
\begin{tabular}{lccccccc}
\toprule
Dataset &
$w_{\ell_1}$ &
$w_{\mathrm{MSE}}$ &
$w_{\mathrm{SSIM}}$ &
$\lambda_x$ &
$\lambda_z$ &
$\lambda_{\mathrm{clean}}$ &
$\lambda_{\mathrm{decor}}$ \\
\midrule
CIFAR-10/100, SVHN
& 1 & 1 & 0.5 & 1 & 0.10 & 0.20 & $10^{-3}$ \\
STL-10, Tiny ImageNet
& 1 & 1 & 0.5 & 1 & 0.05 & 0.05 & $10^{-3}$ \\
\bottomrule
\end{tabular}
\end{table}

\subsection{Dataset and Evaluation Protocol}

\textbf{Datasets and preprocessing: } For CIFAR-10, CIFAR-100, and SVHN, we train on the official training split and use the official test split as the evaluation corpus. For STL-10, we train on the train+unlabeled split and evaluate on the official test split. For Tiny ImageNet, we train on the training set and use the official validation set as the evaluation corpus. CIFAR-10/100 and SVHN are evaluated at $32\times32$ resolution, Tiny ImageNet at $64\times64$, and STL-10 at its native $96\times96$ resolution. Training images use random horizontal flipping, while evaluation uses deterministic preprocessing without augmentation.

\textbf{Corpus-compression setting: } Our primary reconstruction experiments study corpus-specific representation storage rather than conventional held-out autoencoder generalization. We therefore treat the official evaluation split as an unlabeled target corpus whose representations are to be compressed and reconstructed. Under this setting, reconstruction quality on the target corpus is available during model selection, and we retain the checkpoint with the highest reconstruction SSIM. Appendix~\ref{app:protocol_robustness} additionally evaluates a stricter protocol in which the evaluation corpus is completely excluded from checkpoint selection; SLAE retains its matched-storage advantage under this setting.

\textbf{Grouping and evaluation: } For SLAE, examples are randomly partitioned into groups of $K$ and each group shares one superposed memory tensor. Group assignments are fixed within each run, while the order of groups is shuffled during training. Plain AE is evaluated independently on each example, whereas SLAE reconstructs all $K$ examples from their shared memory. We report MSE, SSIM, and PSNR averaged over reconstructed examples. The main experiments use random grouping for all $K\in\{2,4,8\}$.

\section{Downstream Classification Details}
\label{app:downstream_classification}

We evaluate downstream utility on CIFAR-100 using the $H=2$ models from the main experimental grid. For each Plain AE or SLAE configuration, the pretrained compression model is frozen and used to reconstruct the complete CIFAR-100 training set. A CIFAR-style ResNet-18 is then trained from scratch using only these reconstructed images and their labels, and evaluated on the same original clean CIFAR-100 test set. Training and evaluation on the uncompressed data provide the upper reference. Reconstructed images are clipped to $[0,1]$ and materialized as 8-bit images before classifier training.

\begin{table}[H]
\centering
\small
\caption{\textbf{Downstream classification at matched storage.}
CIFAR-100 final-epoch top-1 accuracy (\%). Numbers in parentheses give the
latent channel width $C$.}
\label{tab:downstream_full}
\begin{tabular}{c|c|ccc}
\toprule
$B$ & Plain AE & SLAE $K=2$ & SLAE $K=4$ & SLAE $K=8$ \\
\midrule
32  & 9.28 (8)   & \textbf{21.11} (16) & \textbf{18.08} (32) & \textbf{18.42} (64) \\
64  & 18.42 (16) & \textbf{27.72} (32) & \textbf{35.21} (64) & \textbf{31.42} (128) \\
128 & 27.61 (32) & \textbf{39.53} (64) & \textbf{36.22} (128) & 26.93 (256) \\
256 & 46.36 (64) & \textbf{53.57} (128) & \textbf{49.63} (256) & -- \\
512 & 64.75 (128)& \textbf{67.38} (256) & -- & -- \\
\bottomrule
\end{tabular}
\end{table}

\textbf{Classifier protocol: } All classifiers use the same CIFAR-style ResNet-18 with a $3\times3$ convolutional stem, four residual stages of widths $64$, $128$, $256$, and $512$, global average pooling, and a 100-way linear head. Classifiers are trained for 200 epochs with cross-entropy loss, SGD with Nesterov momentum $0.9$, initial learning rate $0.1$, weight decay $5\times10^{-4}$, batch size 256, and cosine learning-rate decay. Training augmentation consists of four-pixel reflection padding followed by a random $32\times32$ crop and horizontal flipping with probability $0.5$. All runs use the same seed, which is reset immediately before classifier initialization, so initialization, data shuffling, and augmentation randomness are matched across methods. Thus, the only condition-dependent classifier input is the reconstructed training images. We report final-epoch top-1 accuracy after the fixed 200-epoch training in Table~\ref{tab:downstream_full}.

SLAE improves accuracy in 11 of 12 matched-storage comparisons, including all evaluated $K=2$ and $K=4$ settings. The largest gain is $16.79$ percentage points at $B=64$ with $K=4$. The single degradation occurs at the most aggressive evaluated $K=8$ setting, consistent with the capacity--interference tradeoff observed in the reconstruction experiments.

\section{Robustness to Checkpoint-Selection Protocol}
\label{app:protocol_robustness}

Our primary reconstruction experiments follow the corpus-compression setting described in Section~\ref{app:implementation_detail}, where the evaluation(test) split is treated as the unlabeled target corpus whose representations are to be stored. Under this setting, reconstruction quality on the target corpus is available for checkpoint selection. Although this matches the intended corpus-specific storage setting, we additionally test whether the observed SLAE gains depend on access to the evaluation split during model selection.

We therefore repeat representative matched-storage experiments while completely removing the evaluation split from checkpoint selection. Training and model selection use no reconstruction measurements from the evaluation split, which is evaluated only once after training for the final reported result. We otherwise retain the corresponding model configurations and training setup from the main experiments. The selected settings cover two superposition factors on CIFAR-10, and one superposition factor for CIFAR-100 and Tiny ImageNet.

\begin{table}[H]
\centering
\caption{\textbf{Robustness to checkpoint-selection protocol.} We repeat representative matched-storage comparisons while removing the official evaluation split from checkpoint selection. Lower MSE is better; ``Red.'' is the relative MSE reduction of SLAE over Plain.}
\label{tab:protocol_robustness}
\small
\setlength{\tabcolsep}{4pt}
\begin{tabular}{lcccccc}
\toprule
& & & \multicolumn{2}{c}{Original protocol} & \multicolumn{2}{c}{No test selection} \\
\cmidrule(lr){4-5}\cmidrule(lr){6-7}
Dataset & $K$ & $B$ & Plain / SLAE MSE & Red. & Plain / SLAE MSE & Red. \\
\midrule
CIFAR-10 & 2 & 128 & 0.007354 / 0.003933 & 46.5\% & 0.009048 / 0.004046 & 55.3\% \\
CIFAR-10 & 4 & 128 & 0.007354 / 0.004280 & 41.8\% & 0.009048 / 0.004093 & 54.8\% \\
CIFAR-100 & 2 & 128 & 0.007556 / 0.004124 & 45.4\% & 0.008977 / 0.004234 & 52.8\% \\
TinyImageNet & 2 & 512 & 0.008245 / 0.007568 & 8.2\% & 0.007986 / 0.007179 & 10.1\% \\
\bottomrule
\end{tabular}
\end{table}

As shown in Table~\ref{tab:protocol_robustness}, SLAE retains its
matched-storage advantage in all four settings when the evaluation split has no
role in checkpoint selection. The resulting MSE reductions are $55.3\%$ and
$54.8\%$ on CIFAR-10 for $K=2$ and $K=4$, respectively, $52.8\%$ on
CIFAR-100, and $10.1\%$ on Tiny ImageNet. The absolute reconstruction errors
change under the alternative checkpoint-selection protocol, but the relative
advantage of SLAE remains comparable to or larger than under the primary
protocol. In particular, the improvement also persists on Tiny ImageNet, where
the original matched-storage gain is comparatively modest. These results
indicate that the main reconstruction advantage of SLAE is not driven by using
target-corpus reconstruction quality for checkpoint selection.

\section{Full Reconstruction Results}
\label{app:full_reconstruction}

\begin{figure}[H]
    \centering
    \includegraphics[width=\textwidth]{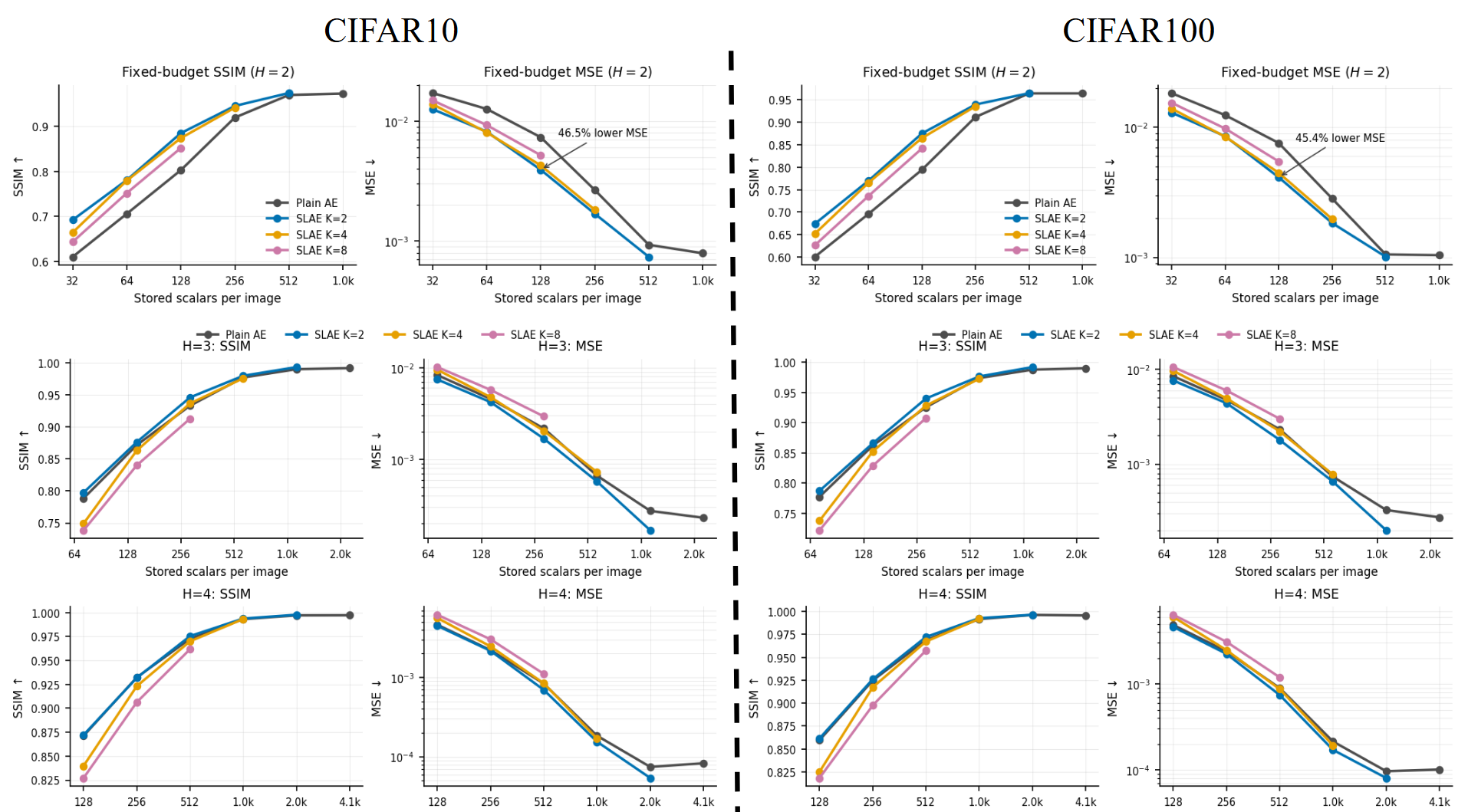}
    \caption{\textbf{Full reconstruction results on CIFAR-10 and CIFAR-100.}
    Fixed-geometry SSIM and MSE curves for $H\in\{2,3,4\}$.
    Plain AE and SLAE are compared at the same average latent-storage budget,
    with SLAE shown for $K\in\{2,4,8\}$.}
    \label{fig:full_cifar}
\end{figure}

\begin{figure}[H]
    \centering
    \includegraphics[width=\textwidth]{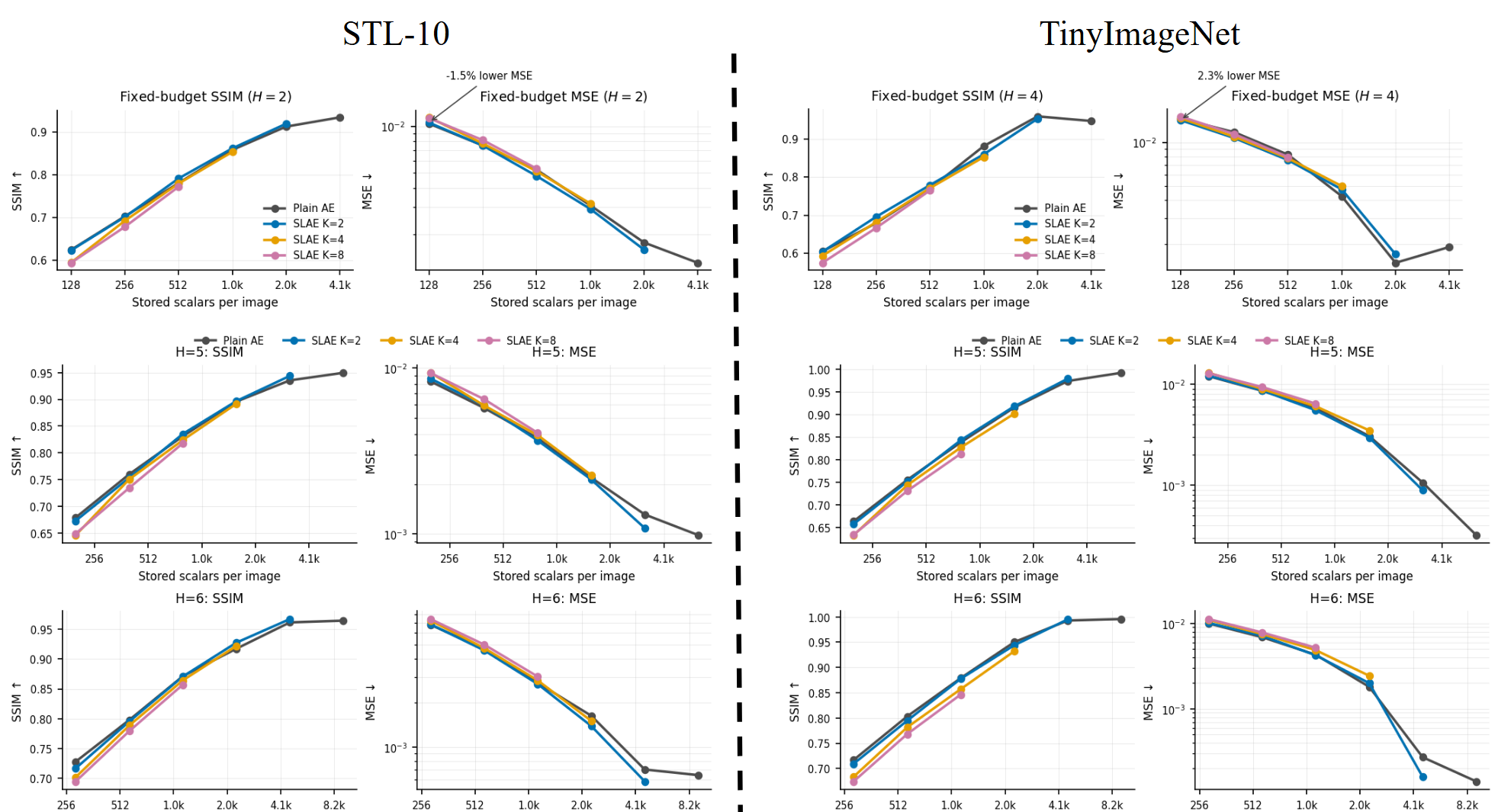}
    \caption{\textbf{Full reconstruction results on STL-10 and Tiny ImageNet.}
    Fixed-geometry SSIM and MSE curves for $H\in\{4,5,6\}$.
    Plain AE and SLAE are compared at matched average latent storage for
    $K\in\{2,4,8\}$.}
    \label{fig:full_stl_tiny}
\end{figure}

\begin{figure}[H]
    \centering
    \includegraphics[width=0.5\linewidth]{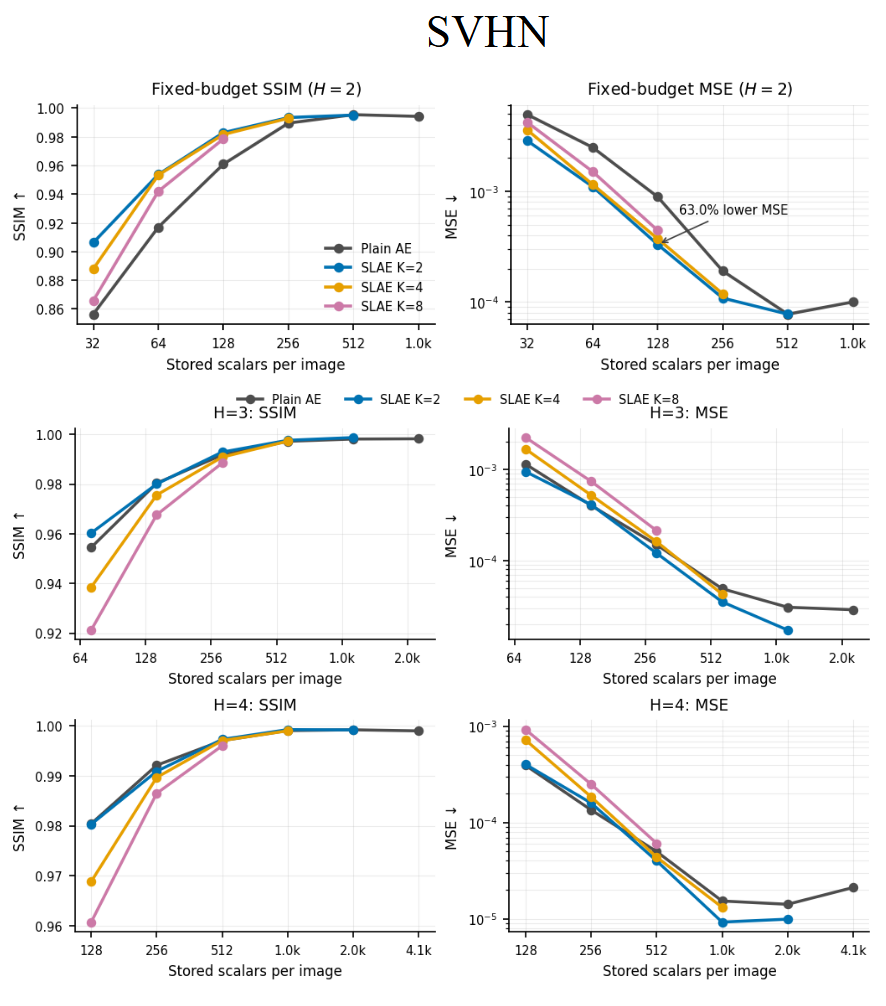}
    \caption{\textbf{Full reconstruction results on SVHN.}
    Fixed-geometry SSIM and MSE curves for $H\in\{2,3,4\}$ at matched average latent storage for $K\in\{2,4,8\}$.}
    \label{fig:full_svhn}
\end{figure}

Figures~\ref{fig:full_cifar}, \ref{fig:full_stl_tiny}, and \ref{fig:full_svhn} report the complete fixed-geometry reconstruction results underlying the summary in Section~\ref{sec:experiments}. For each latent spatial size $H$, we compare Plain AE with SLAE for $K\in\{2,4,8\}$ as a function of the average number of stored scalars per image. All comparisons are storage-matched. We report both SSIM and MSE, with higher SSIM and lower MSE indicating better reconstruction.

Across datasets, moderate superposition ($K=2$) is generally the most reliable operating point, while increasing $K$ provides additional latent capacity at the cost of increasingly difficult recovery. These trends are consistent with the capacity--interference tradeoff analyzed in Section~5.

\section{Additional Capacity--Interference Results}
\label{app:capacity_interference}

The analysis in Section~\ref{sec:analysis} summarizes the tradeoff between the benefit of a wider latent representation and the penalty introduced by superposition. Here we show the same decomposition separately for each dataset. Recall that, at matched per-example storage $B$,
\begin{equation}
G_{\mathrm{cap}} = \mathcal{E}_B-\mathcal{E}_{KB}
\end{equation}
measures the reconstruction improvement obtained by replacing the narrow Plain AE with a $K$-times-wider latent, while
\begin{equation}
P_{\mathrm{sup}} = \mathcal{E}_{\mathrm{SLAE}}-\mathcal{E}_{KB}
\end{equation}
measures how much of this gain is lost through superposition and recovery. Their difference is exactly the SLAE improvement over the matched-storage Plain AE:
\begin{equation}
G_{\mathrm{cap}}-P_{\mathrm{sup}}
=
\mathcal{E}_B-\mathcal{E}_{\mathrm{SLAE}}
\end{equation}
Consequently, SLAE is beneficial precisely when
\begin{equation}
P_{\mathrm{sup}} < G_{\mathrm{cap}}
\end{equation}

\begin{figure}[H]
    \centering
    \includegraphics[width=\textwidth]{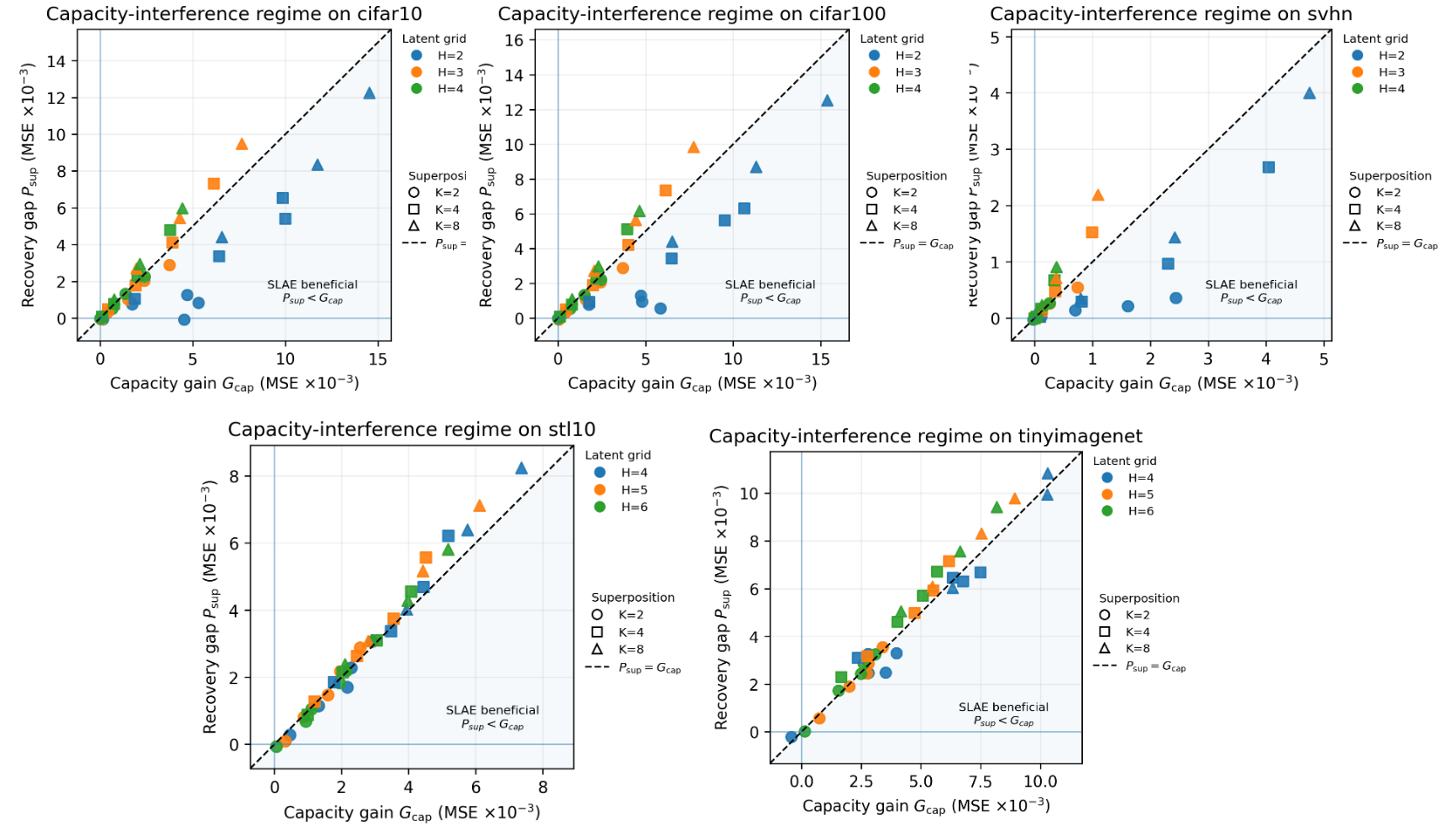}
    \caption{\textbf{Dataset-wise capacity--interference decomposition.}
    Each point represents a matched-storage operating point. The horizontal axis
    shows the capacity gain $G_{\mathrm{cap}}$ obtained from the corresponding
    $K$-times-wider Plain AE, and the vertical axis shows the recovery penalty
    $P_{\mathrm{sup}}$ incurred by SLAE relative to that wide representation.
    Points below the dashed line $P_{\mathrm{sup}}=G_{\mathrm{cap}}$ correspond
    exactly to settings in which SLAE improves over the matched-storage Plain AE.
    Colors indicate latent spatial size $H$, while marker shapes indicate
    $K\in\{2,4,8\}$.}
    \label{fig:capacity_interference_datasets}
\end{figure}

Figure~\ref{fig:capacity_interference_datasets} shows that SLAE works best when the benefit of using a wider latent representation is larger than the cost introduced by superposition. This pattern is especially clear on CIFAR-10, CIFAR-100, and SVHN, while STL-10 and Tiny ImageNet show a more balanced tradeoff. As $K$ increases, recovery becomes harder and the interference cost grows, making the benefit of superposition less consistent. Overall, these results support the central intuition of SLAE: superposition is most useful when the extra latent capacity provides enough benefit to outweigh the interference introduced by storing multiple representations together.

\section{Derivation of Randomized-Binding Interference}
\label{app:binding_derivation}

This section derives the directional interference moments stated in equations~\ref {eq:interference}--~\ref{eq:interference_std}. We condition on fixed storage codes that are independent of the randomized binding keys, and take expectations only over independently sampled slot keys.

\textbf{Binding transform: } At each spatial location, let the storage-code channel dimension be $d=KC$. SLAE binds the code in slot $i$ using
\begin{equation}
B_i=\mathcal{H}D_iP_i
\end{equation}
where $P_i$ is a uniformly sampled permutation matrix, $D_i$ is diagonal with independent Rademacher entries, and $\mathcal{H}$ is the normalized
Walsh--Hadamard transform. Each factor is orthogonal, so $B_i^{-1}=B_i^\top$. For a shared memory
\begin{equation}
m=\frac{1}{\sqrt{K}}\sum_{i=1}^{K}B_i s_i
\end{equation}
retrieving slot $k$ gives
\begin{align}
\sqrt{K}B_k^\top m
&=
\sum_{i=1}^{K} B_k^\top B_i s_i \nonumber\\
&=
s_k+\sum_{i\neq k}Q_{ki}s_i
\label{eq:app_unbind}
\end{align}
where
\begin{equation}
Q_{ki}
=
B_k^\top B_i
=
P_k^\top D_k\mathcal{H}^\top\mathcal{H}D_iP_i
=
P_k^\top D_kD_iP_i
\end{equation}
Thus, the shared Hadamard transform cancels in the relative transform, and $Q_{ki}$ is itself a randomized signed permutation. At one spatial position, let $v\in\mathbb{R}^d$ denote the channel vector of an interfering storage code and let $a\in\mathbb{R}^d$ be any fixed direction, such directions can be viewed as local linear features used by subsequent recovery or decoding layers. The scalar interference contributed along $a$ is $a^\top Qv$.

\textbf{Directional interference from one slot: } A randomized signed permutation can be written componentwise as
\begin{equation}
(Qv)_j=\sigma_j v_{\pi(j)}
\end{equation}
where $\pi$ is a uniformly random permutation of $\{1,\ldots,d\}$ and $\sigma_1,\ldots,\sigma_d$ are independent Rademacher variables. Hence
\begin{equation}
a^\top Qv
=
\sum_{j=1}^{d}a_j\sigma_jv_{\pi(j)}
\end{equation}
Conditioning on $\pi$ and averaging over the random signs,
\begin{equation}
\mathbb{E}_{\sigma}
\left[a^\top Qv\mid\pi\right]
=
\sum_{j=1}^{d}
a_jv_{\pi(j)}\mathbb{E}[\sigma_j]
=
0
\end{equation}
and therefore
\begin{equation}
\mathbb{E}_{r}[a^\top Qv]=0
\end{equation}

For the second moment,
\begin{align}
\mathbb{E}_{\sigma}
\left[(a^\top Qv)^2\mid\pi\right]
&=
\mathbb{E}_{\sigma}
\left[
\sum_{j,\ell}
a_ja_\ell
\sigma_j\sigma_\ell
v_{\pi(j)}v_{\pi(\ell)}
\right]
\nonumber\\
&=
\sum_{j=1}^{d}a_j^2v_{\pi(j)}^2
\label{eq:app_second_cond}
\end{align}
because
$\mathbb{E}[\sigma_j\sigma_\ell]=0$ for $j\neq\ell$ and equals $1$ for $j=\ell$. Since $\pi(j)$ is uniformly distributed over $\{1,\ldots,d\}$,
\begin{equation}
\mathbb{E}_{\pi}
\left[v_{\pi(j)}^2\right]
=
\frac{1}{d}\sum_{\ell=1}^{d}v_\ell^2
=
\frac{\|v\|_2^2}{d}
\end{equation}
Averaging equation~\ref{eq:app_second_cond} over $\pi$ therefore yields
\begin{equation}
\mathbb{E}_{r}\left[(a^\top Qv)^2\right]
=
\frac{\|a\|_2^2\|v\|_2^2}{d}
\end{equation}
Together with the zero-mean result above, this proves equation~\ref{eq:interference}.

\textbf{Directional interference from multiple slots: } For retrieval of slot $k$, the total interference along direction $a$ is
\begin{equation}
\epsilon_k
=
\sum_{i\neq k}a^\top Q_{ki}v_i
\end{equation}
Each term has zero mean by equation~\ref{eq:interference}, so
\begin{equation}
\mathbb{E}_{r}[\epsilon_k]=0
\end{equation}

Although the relative transforms $Q_{ki}$ share the target key $B_k$, the variance can be computed by conditioning on that key. Given $B_k$, the keys $\{B_i\}_{i\neq k}$ are independent, and the corresponding interference terms are conditionally independent with zero conditional mean. Hence, for $i\neq j$,
\begin{equation}
\mathbb{E}_{r}
\left[
(a^\top Q_{ki}v_i)
(a^\top Q_{kj}v_j)
\right]
=0
\end{equation}
The variances therefore add, giving
\begin{equation}
\operatorname{Var}_{r}(\epsilon_k)
=
\frac{\|a\|_2^2}{d}
\sum_{i\neq k}\|v_i\|_2^2
\end{equation}
Taking the square root gives
\begin{equation}
\operatorname{Std}_{r}(\epsilon_k)
=
\frac{\|a\|_2}{\sqrt d}
\left(
\sum_{i\neq k}\|v_i\|_2^2
\right)^{1/2}
\end{equation}
which proves equation~\ref{eq:interference_std}. In particular, if $\|a\|_2=1$ and the interfering codes have comparable norm $\|v_i\|_2\approx\rho$, then
\begin{equation}
\operatorname{Std}_{r}(\epsilon_k)
\approx
\rho\sqrt{\frac{K-1}{d}}
\end{equation}

\paragraph{Directional dispersion versus total interference energy: } The $1/\sqrt d$ scaling above concerns the projection of interference onto a fixed direction; it does not imply that the complete interference vector has small norm. Let
\begin{equation}
\eta_k
=
\sum_{i\neq k}Q_{ki}s_i
\end{equation}
Since every $Q_{ki}$ is orthogonal,
\begin{equation}
\|Q_{ki}s_i\|_F=\|s_i\|_F
\end{equation}
The expected cross terms between distinct interfering slots vanish under the independent random-key model, and therefore
\begin{equation}
\mathbb{E}_{r}\|\eta_k\|_F^2
=
\sum_{i\neq k}\|s_i\|_F^2
\end{equation}
Randomized binding therefore does not eliminate interference energy. Instead, it randomizes its orientation so that its contribution along any fixed direction is zero-mean and dispersed over the $d$ channel dimensions.

\textbf{Interpretation: } Randomized binding need not preserve every coordinate of the original code. What matters is that subsequent layers can still recover useful linear features. For a readout direction $a$,
\begin{equation}
a^\top(v_k+\eta_k)=a^\top v_k+a^\top\eta_k
\end{equation}
and for fixed code energy, the interference term is zero-mean with standard deviation decreasing as $1/\sqrt d$. Thus, although the full interference vector can remain large, high-dimensional random binding makes it nearly orthogonal to any fixed feature direction, approximately preserving the inner products used by downstream layers.

The analysis above gives intuition to latent superposition and characterizes the randomized binding and unbinding stage for fixed storage codes. It does not imply exact recovery or characterize the nonlinear behavior of the learned storage adapter and recovery network after training. These effects are captured empirically by the recovery penalty analyzed in Sections~\ref{sec:analysis} and~\ref{sec:experiments}.

\section{Memory Savings at Matched Quality}
\label{app:matched_quality}

The main experiments compare reconstruction quality at matched storage. Here we consider the complementary question: {at a fixed target reconstruction quality, how much storage is required?} For a target quality $q$ and spatial latent size $H$, let $B_{\mathrm{Plain}}(q,H)$ and $B_{\mathrm{SLAE}}(q,H)$ denote the smallest evaluated storage budgets at which Plain AE and SLAE, respectively, reach the target. We report the memory reduction factor
\begin{equation}
R(q,H)
=
\frac{B_{\mathrm{Plain}}(q,H)}
     {B_{\mathrm{SLAE}}(q,H)}
\end{equation}
Thus, $R=2$ means that SLAE reaches the same target quality using half as many stored scalars, $R=1$ indicates equal storage, and $R<1$ indicates that SLAE requires more storage. We use representative dataset-specific SSIM and MSE targets within the informative range of each dataset's reconstruction curves. Targets that are not reached within the evaluated storage range are marked as not reached.

\begin{figure}[H]
    \centering
    \includegraphics[width=\textwidth]{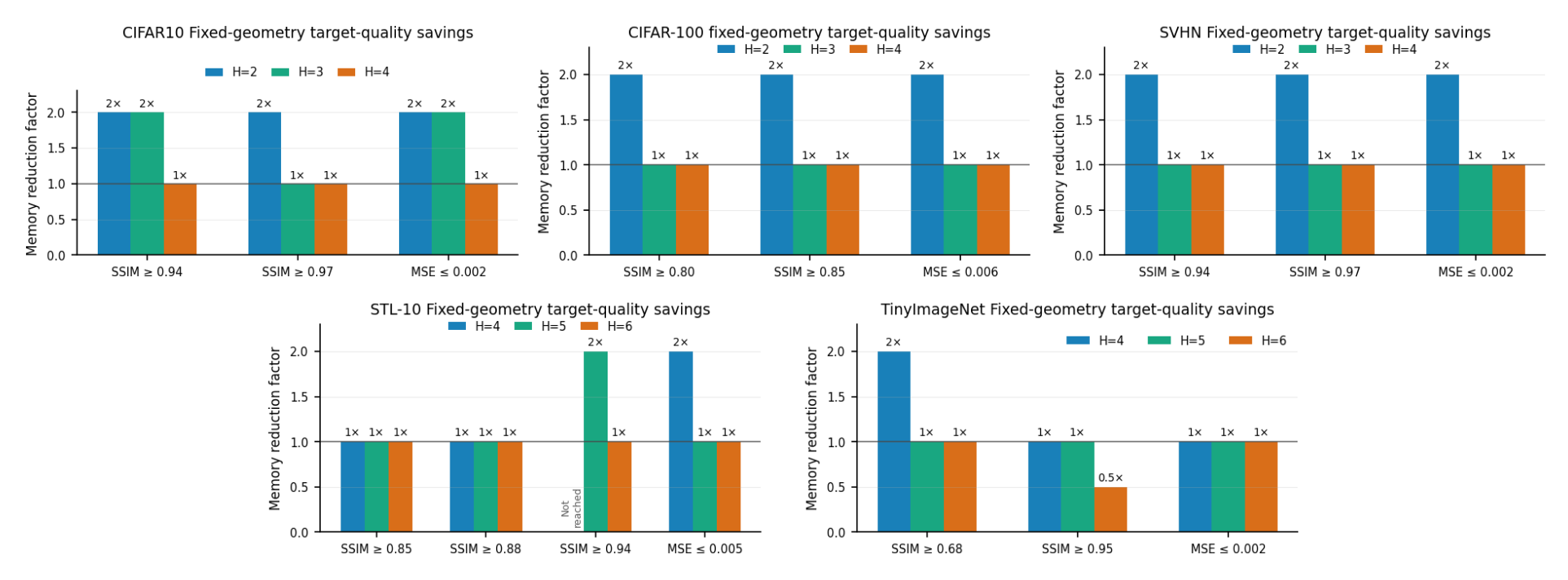}
    \caption{\textbf{Fixed-geometry memory efficiency at matched reconstruction quality.}
    For each dataset and latent spatial size $H$, we report the ratio of the minimum evaluated Plain AE storage budget to the minimum evaluated SLAE storage budget required to reach each target quality. Values above $1\times$ indicate an SLAE memory saving, $1\times$ indicates parity, and values below $1\times$ indicate that SLAE requires more storage. Quality targets are selected separately for each dataset to reflect its observed reconstruction range.}
    \label{fig:matched_quality_savings}
\end{figure}

Figure~\ref{fig:matched_quality_savings} provides a complementary view of the matched-storage results. On CIFAR-10, CIFAR-100, and SVHN, SLAE reaches several representative quality targets with a $2\times$ reduction in storage in the most constrained latent geometry. The savings generally diminish as $H$ increases and the conventional autoencoder becomes less capacity limited. STL-10 and Tiny ImageNet show a more mixed pattern, with savings at selected operating points but mostly parity elsewhere. This is consistent with the capacity--interference picture in the main paper: shared storage is most beneficial when additional latent capacity is valuable enough to outweigh the recovery cost introduced by superposition.

\section{Comparison with Alternative Latent Compression Strategies}
\label{app:alternative_compression}

The main experiments compare SLAE against conventional autoencoders under matched fixed-precision latent storage. Here we ask a broader question: how does superposed storage compare with, and combine with, alternative mechanisms for reducing latent-memory cost? We consider representative dimensional-compression approaches, including lower-dimensional autoencoder bottlenecks \citep{allerbo2021non,hinton2006reducing} and PCA \citep{baldi1989neural}, as well as numerical precision reduction through post-training quantization \citep{wei2022qdrop,liu2023pd}. Recent learned compression systems similarly combine compact latent representations with quantization to trade storage rate against reconstruction quality \citep{liu2023learned}.

We evaluate these alternatives on CIFAR-10 in the controlled $H=2$ setting, using the $C=64$ Wide Plain latent as the uncompressed reference. This latent contains $64\times2\times2=256$ values, corresponding to $8192$ bits per image in FP32. We consider compression ratios of $2\times$, $4\times$, and $8\times$ relative to this reference.

\textbf{Baselines and bit accounting:}
We compare four storage strategies. Reduced AE decreases the latent width to $C\in\{32,16,8\}$ while retaining FP32 storage. PCA projects the Wide Plain $C=64$ latent to $128$, $64$, or $32$ FP32 coefficients, with the PCA basis fitted using training-set latents. Quantization retains the full $C=64$ latent but stores its values using FP16, INT8, or INT4. FP16 is implemented as a precision-cast round trip, while INT8 and INT4 use post-training per-channel uniform quantization with scales calibrated on the training set. Finally, SLAE retains the $C=64$ latent width and uses $K\in\{2,4,8\}$-way superposition. We additionally apply precision reduction to the shared SLAE memory after superposition, yielding combined configurations such as $K=2+$FP16 and $K=2+$INT8. For a shared memory stored using $b$ bits per value, the effective compression relative to independently stored FP32 wide latents is therefore $\rho = K\frac{32}{b}$.

\begin{figure}[H]
    \centering
    \includegraphics[width=\textwidth]{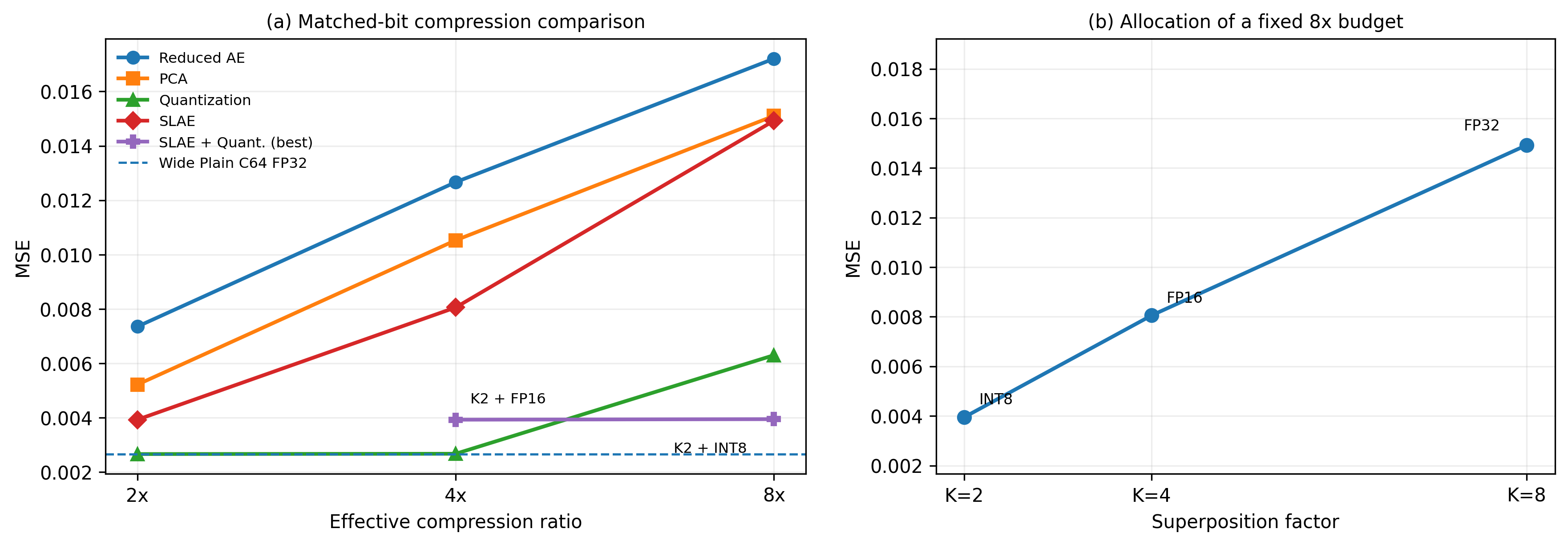}
    \caption{\textbf{Alternative latent-compression strategies on CIFAR-10.} We use $H=2$ and a $C=64$ Wide Plain FP32 latent as the reference. \textbf{(a)} Reconstruction MSE under matched per-image latent-memory bit budgets. SLAE outperforms dimensional compression through both reduced-width autoencoders and PCA, while numerical precision reduction is highly effective at mild compression. At $8\times$, combining $K=2$ superposition with INT8 shared-memory storage achieves the lowest MSE with $37.3\%$ reduction relative to the strongest INT4 quantization at the same bit budget. \textbf{(b)} Allocation of a fixed $8\times$ compression budget between superposition and numerical precision. Moderate $K=2$ superposition with INT8 storage substantially outperforms allocating more of the compression to superposition.}
    \label{fig:alternative_compression}
\end{figure}

\textbf{Superposition versus dimensional compression: } Figure~\ref{fig:alternative_compression}(a) and Table~\ref{tab:alternative_compression} first show that SLAE provides a clear advantage over methods that reduce latent dimensionality. At $2\times$ compression, $K=2$ SLAE obtains MSE $0.00393$, compared with $0.00522$ for PCA and $0.00735$ for the reduced-width AE, corresponding to MSE reductions of $24.7\%$ and $46.6\%$, respectively. The same pattern persists at $4\times$: $K=4$ SLAE achieves MSE $0.00806$, compared with $0.01053$ for PCA and $0.01267$ for the reduced-width AE. These results reinforce the distinction underlying SLAE: preserving a wider representation and tolerating recoverable interference can retain substantially more information than irreversibly reducing latent dimensionality.

\textbf{Superposition is complementary to precision compression: } Precision reduction is particularly effective at mild compression. FP16 storage at $2\times$ is effectively indistinguishable from the FP32 Wide Plain reference, and INT8 at $4\times$ similarly introduces negligible reconstruction degradation. Thus, SLAE should not be viewed as a replacement for numerical quantization; the two mechanisms reduce memory along different axes. Quantization reduces the bits used by each stored value, whereas SLAE reduces the number of independently stored representations by allowing multiple examples to share one memory tensor.

This distinction becomes important at more aggressive compression. Achieving $8\times$ compression through precision reduction alone requires INT4 storage, which increases MSE to $0.00630$. In contrast, combining $K=2$ SLAE with INT8 shared-memory storage achieves MSE $0.00395$, SSIM $0.88461$, and PSNR $24.04$~dB at the same $1024$-bit-per-image budget. This corresponds to a $37.3\%$ reduction in MSE relative to direct INT4 quantization. Moreover, quantizing the $K=2$ shared memory from FP32 to INT8 changes MSE only from $0.00393$ to $0.00395$, indicating that the learned superposed memory remains robust to moderate post-training quantization.

Figure~\ref{fig:alternative_compression}(b) further isolates how a fixed $8\times$ budget should be divided between the two mechanisms. The three equal-budget configurations obtain
$$
    0.00395\;(K=2+\mathrm{INT8}),\qquad
    0.00806\;(K=4+\mathrm{FP16}),\qquad
    0.01493\;(K=8+\mathrm{FP32})
$$
Performance therefore degrades as increasingly more of the compression burden is assigned to superposition. This is consistent with the capacity--interference analysis in the main paper: larger $K$ provides additional latent capacity, but also introduces greater recovery difficulty. Quantization provides a complementary way to reduce storage without increasing cross-slot interference, making moderate superposition combined with moderate quantization particularly effective.

\begin{table}[H]
\centering
\caption{Alternative latent-compression strategies on CIFAR-10 with $H=2$ and a $C=64$ wide reference latent. Methods are compared at matched per-image latent-memory bit budgets.}
\label{tab:alternative_compression}
\begin{tabular}{clrrrr}
\toprule
Compression & Method & Bits/img & MSE $\downarrow$ & SSIM $\uparrow$ & PSNR $\uparrow$ \\
\midrule
$1\times$ & Wide Plain C64 FP32 & 8192 & \textbf{0.002666} & \textbf{0.91973} & \textbf{25.745} \\
\midrule
$2\times$ & Reduced AE C32 FP32 & 4096 & 0.007354 & 0.80276 & 21.338 \\
 & PCA-128 FP32 & 4096 & 0.005220 & 0.84858 & 22.827 \\
 & Quant C64 FP16 & 4096 & \textbf{0.002666} & \textbf{0.91973} & \textbf{25.745} \\
 & SLAE K2 FP32 & 4096 & 0.003930& 0.88500& 24.058\\
\midrule
$4\times$ & Reduced AE C16 FP32 & 2048 & 0.012665 & 0.70591 & 18.976 \\
 & PCA-64 FP32 & 2048 & 0.010528 & 0.72199 & 19.779 \\
 & Quant C64 INT8 & 2048 & \textbf{0.002678} & \textbf{0.91946} & \textbf{25.726} \\
 & SLAE K4 FP32 & 2048 & 0.008058& 0.77968& 20.939\\
 & SLAE K2 + FP16 & 2048 & 0.003930 & 0.88500 & 24.058 \\
\midrule
$8\times$ & Reduced AE C8 FP32 & 1024 & 0.017206 & 0.61033 & 17.645 \\
 & PCA-32 FP32 & 1024 & 0.015104 & 0.61783 & 18.211 \\
 & Quant C64 INT4 & 1024 & 0.006302 & 0.84138 & 22.006 \\
 & SLAE K8 FP32 & 1024 & 0.014925 & 0.64356& 18.261 \\
 & SLAE K2 + INT8 & 1024 & \textbf{0.003950} & \textbf{0.88461} & \textbf{24.037} \\
 & SLAE K4 + FP16 & 1024 & 0.008058 & 0.77968& 20.939\\
\bottomrule
\end{tabular}
\end{table}


\section{Limitations}
\label{app:limitations}

While SLAE demonstrates a strong reconstruction--memory tradeoff across a range of datasets and storage regimes, the current approach has several limitations that define important directions for future work.
\begin{itemize}
    \item \textbf{High-$K$ regime: } More aggressive superposition provides wider latent representations but also increases the recovery penalty. In our experiments, $K=2$ is the most reliable operating point, while $K=4$ and $K=8$ are beneficial only in selected regimes.

    \item \textbf{Parameter overhead at very wide latents: } The storage adapter and recovery network introduce modest overhead in the capacity-limited regimes where SLAE performs best, but the current recovery architecture becomes parameter-expensive at very large channel widths.

    \item \textbf{Fixed superposition factor: } Each value of $K$ is trained separately, so the current model cannot dynamically change the degree of superposition at deployment time. Supporting variable $K$ within a single model would enable more flexible memory allocation.
\end{itemize}

Overall, these limitations motivate future work on more scalable recovery, adaptive superposition, and broader forms of learned representation storage.

\section{Discussion}
\label{app:discussion}

Our results suggest several broader directions for superposed representation storage beyond the specific SLAE design studied in this work.

\textbf{Superposition as a complementary compression axis: } Our results suggest that superposition should be viewed not as a replacement for existing compression techniques, but as an additional axis along which representation memory can be reduced. Conventional methods primarily compress within each representation, for example by reducing dimensionality, numerical precision, or coding redundancy. SLAE instead compresses across representations by allowing multiple high-capacity latents to share physical storage. Because these mechanisms act on different sources of memory cost, they can therefore be composed. The results in Appendix~\ref{app:alternative_compression} provide an initial demonstration of this direction, showing that moderate superposition combined with quantization can yield a better memory--quality tradeoff than relying aggressively on either mechanism alone. More broadly, this suggests a path toward multi-axis representation compression systems that jointly optimize latent capacity, superposition, quantization, and coding efficiency under a shared memory budget.

\textbf{Alternative binding designs: } We use a fixed randomized orthogonal binding scheme based on signed channel permutations and a shared Walsh--Hadamard transform. This is only one possible design. Learned keys, alternative orthogonal transforms, spatial--channel binding, or multiple independent binding measurements may provide better interference separation or easier recovery, particularly at larger $K$.

\textbf{Better grouping strategies: } The current experiments randomly group examples before superposition. However, the recoverability of a group may depend on the composition of its members. A memory system could instead learn or select groups to minimize interference, potentially placing together representations whose storage codes are especially compatible under superposition.

\textbf{Beyond image reconstruction: } SLAE is not inherently tied to image autoencoders. The same principle could be studied for continual-learning replay buffers, stored neural features, multimodal or language-model representations, and vector-memory systems. In these settings, the relevant objective may be downstream task performance or retrieval fidelity rather than pixel-level reconstruction, potentially creating different capacity--interference operating regimes.

\end{document}